\documentclass[journal]{IEEEtran}
\usepackage[T1]{fontenc}
\usepackage{amsmath,amssymb,amsfonts}
\usepackage{algorithmic}
\usepackage{algorithm}
\usepackage{graphicx}
\usepackage{wrapfig}
\usepackage{textcomp}
\usepackage{xcolor}
\usepackage{booktabs}
\usepackage{multirow}
\usepackage{makecell}
\usepackage{placeins}
\usepackage{subcaption}
\usepackage{hyperref}
\usepackage{cleveref}

\begin{document}

\title{CreatiParser: Generative Image Parsing of Raster Graphic Designs into Editable Layers}

\author{Weidong Chen,~\IEEEmembership{Member,~IEEE},
        Dexiang Hong, Zhendong Mao,~\IEEEmembership{Member,~IEEE},
        
        Yutao Cheng,
        Xinyan Liu,
        Lei Zhang,
        Yongdong Zhang,~\IEEEmembership{Fellow,~IEEE}
\thanks{This research is supported by Artificial Intelligence-National Science and Technology Major Project 2023ZD0121200, the National Natural Science Foundation of China under Grant 62302474 and 62502115. Corresponding author: Zhendong Mao.}
\thanks{Weidong Chen, Dexiang Hong, and Lei Zhang are with the School of Information Science and Technology, University of Science and Technology of China, Hefei 230027, China (e-mail: chenweidong@ustc.edu.cn; hongdexiang@mail.ustc.edu.cn; leizh23@ustc.edu.cn).}
\thanks{Yutao Cheng is Individual Researcher (e-mail: taorebobi@gmail.com).}
\thanks{Xinyan Liu is with the School of Computer Science and Technology, Harbin Institute of Technology (Weihai), Weihai, China (e-mail: xinyliu@hit.edu.cn).}
\thanks{Zhendong Mao and Yongdong Zhang, are with the School of Information Science and Technology, University of Science and Technology of China, and are also with the Institute of Artificial Intelligence, Hefei Comprehensive National Science Center, Hefei 230027, China (e-mail: zdmao@ustc.edu.cn; zhyd73@ustc.edu.cn).}}

\maketitle

\begin{abstract}
Graphic design images consist of multiple editable layers, such as text, background, and decorative elements, while most generative models produce rasterized outputs without explicit layer structures, limiting downstream editing. Existing graphic design parsing methods typically rely on multi-stage pipelines combining layout prediction, matting, and inpainting, which suffer from error accumulation and limited controllability. We propose a hybrid generative framework for raster-to-layer graphic design parsing that decomposes a design image into editable text, background, and sticker layers. Text regions are parsed using a vision-language model into a text rendering protocol, enabling faithful reconstruction and flexible re-editing, while background and sticker layers are generated using a multi-branch diffusion architecture with RGBA support. We further introduce ParserReward and integrate it with Group Relative Policy Optimization to align generation quality with human design preferences. Extensive experiments on two challenging datasets, \emph{i.e.,} the Parser-40K and Crello datasets, demonstrate superior performance over existing methods, \emph{eg.,} achieving an overall average improvement of 23.7\% across all metrics.\footnote{Code and the Parser-40K dataset will be released in the final version of the paper.}
\end{abstract}


\section{Introduction}
Graphic design serves as an indispensable medium for visual communication across industries—from advertising (billboards, social media creatives) to UI/UX design (app interfaces, web banners). A defining characteristic of graphic design is its layered composition, typically comprising text, stickers, and backgrounds. The existence of editable layers allows designers to flexibly adjust individual components: for instance, updating promotional text on a poster without altering the background texture, or resizing a product image in a banner while maintaining overall color harmony and layout consistency. However, the rapid proliferation of AI-generated raster designs, produced by advanced generative models~\cite{seedream40, cai2025z, xu2024cookgalip, jiang2025mgdefect,yuan2023semantic,pan2024few, zhang2025creatidesign, zhang2025creatilayout, zhang2025creatiposter}, has created a pressing demand for “raster-to-layer” parsing. These AI outputs are pixel-based and non-editable, making them unsuitable for post-generation customization. Consequently, designers often have to rely on manual editing tools (e.g., Photoshop, Adobe Firefly) to separate and reconstruct layers, a process that is both time-consuming and error-prone.

\begin{figure}[t]
  \centering
  \includegraphics[width=\linewidth]{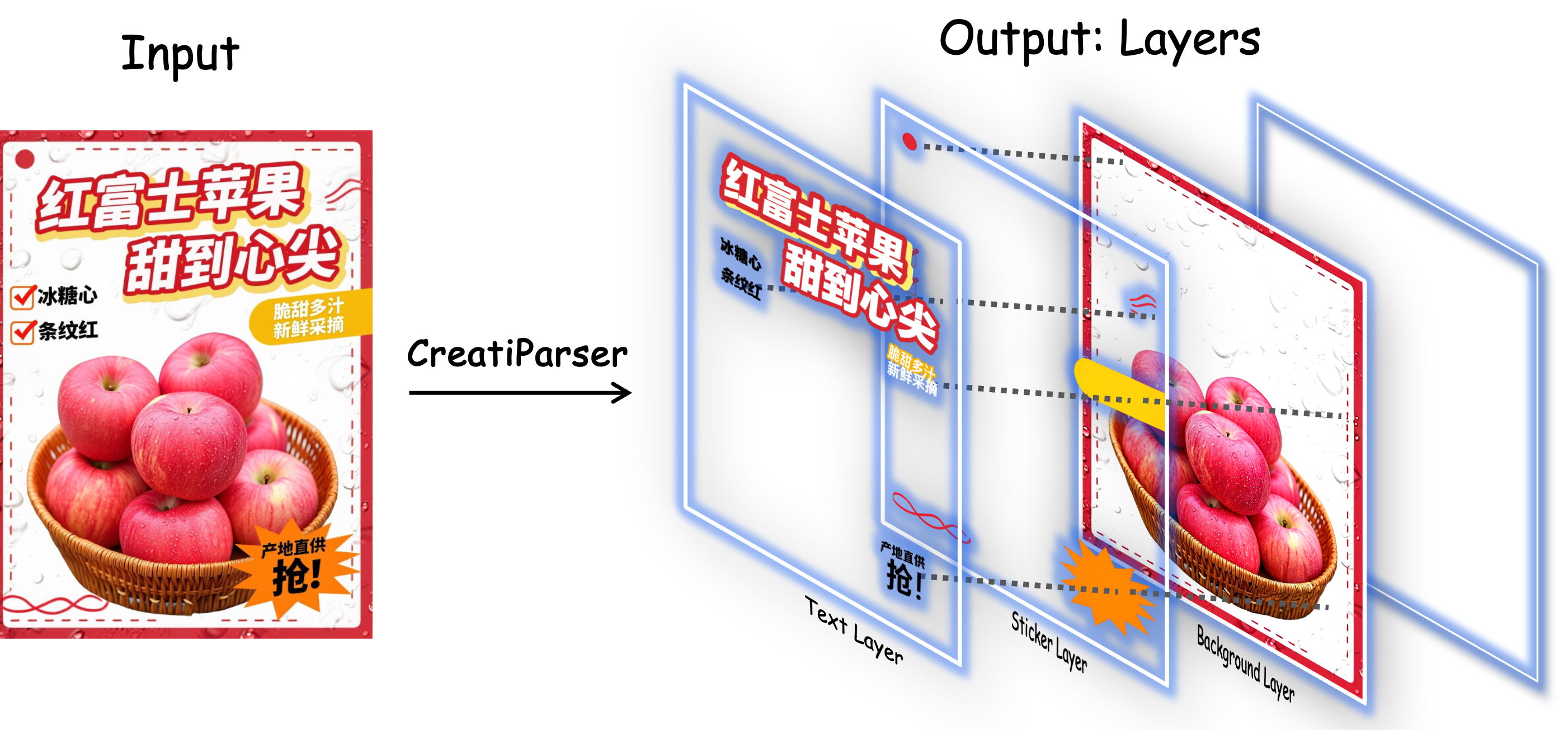}
  \caption{Illustration of graphic design image parsing. Given a rasterized design image, our task aims to decompose it into three editable layers: \textit{Text}, \textit{Sticker}, and \textit{Background}.}
  \label{fig:instruct}
  \vspace{-10pt}
\end{figure}

Recent advances in image parsing have attempted to automate this decomposition, and existing methods can be broadly divided into two categories.
The first category, which dominates the current landscape, leverages multimodal large language models (MLLMs) to parse an image into a draft protocol—a structured intermediate representation specifying the spatial locations of key elements and textual attributes in a predefined format~\cite{niedecomposition, chen2025rethinking, layerd}. Based on this protocol, a cascaded pipeline of first matting and then image inpainting is employed to separate detected elements and reconstruct the background. However, these multi-step modularized systems are prone to error accumulation: inaccuracies in layout parsing propagate to matting, and the uncontrollable behavior of inpainting models~\cite{suvorov2022resolution, zhang2024mutualinpaint} often introduces visual artifacts or stylistic inconsistencies, yielding results that deviate from the original design semantics. 
The second category of methods utilizes generative models for parsing. Prior studies in generative direction, however, have been largely restricted to natural image decomposition, focusing on foreground-background separation~\cite{huang2025dreamlayer,mulan,yang2024generative}. Such a binary paradigm is ill-suited for graphic design, where compositions comprise semantically
heterogeneous elements---such as backgrounds, foreground subjects, and text overlays---serving distinct functions. Collapsing them into a coarse dichotomy forces the model to conflate fundamentally different components, yielding semantically ambiguous layers with limited editability.

In this work, we propose a hybrid generative framework for raster-to-layer graphic design parsing that explicitly models the heterogeneous nature of design elements. Instead of relying on a single generative paradigm, we decompose the problem into text and non-text components. Text regions are parsed using a Vision Language Model, which converts rasterized text into a text rendering protocol, enabling faithful reconstruction and flexible re-editing regardless of font availability or artistic styles. For non-textual content, we design a multi-branch diffusion architecture that jointly generates background and sticker layers---i.e., decorative non-text foreground elements such as lines, shapes, and icons---with explicit support for transparent RGBA outputs.To further align text parsing with human design preferences, we introduce ParserReward, a task-specific evaluation objective that captures semantic fidelity, layer disentanglement, and editability. We integrate ParserReward into a Group Relative Policy Optimization (GRPO) framework to optimize Qwen3-VL text rendering protocol prediction in a stable and parameter-efficient manner. Extensive experiments on various benchmarks demonstrate that our approach consistently outperforms existing methods in terms of layer reconstruction accuracy, text editability, and perceptual quality. Notably, the proposed framework generalizes well to the Crello dataset in a zero-shot setting, highlighting its robustness to unseen design styles and layouts. To sum up, our contributions in this paper can be summarized as follows:

- We propose a hybrid generative framework for raster-to-layer graphic design image parsing, which decomposes a rasterized design image into text layer, background layer, and sticker layer, overcoming the limitations of multi-stage pipeline-based existing methods.

- We introduce a VLM-based text parsing module to predict the text layer, together with a multi-branch diffusion architecture for background and sticker layer generation, enabling faithful reconstruction and flexible editing.

- We propose Parser Reward and integrate it with Group Relative Policy Optimization to align Qwen3-VL text rendering protocol prediction with human design preferences, achieving superior  text layer parsing performance.

- We conduct extensive experiments on various graphic design benchmarks, and our method achieves state-of-the-art performance in terms of layer reconstruction accuracy, text editability, and perceptual quality, demonstrating the effectiveness and generalizability of the proposed framework.

\section{Related Work}
\subsection{Image Decomposition}

In the early stage, image decomposition methods were dominated by optimization with geometric or physical priors in color space~\cite{Tan2016RGBLayer, Tan2018PaletteRGBXY, Aksoy2017Unmixing, Akimoto2020FastSoftSeg, Koyama2018AdvancedBlending, Horita2022Unblending}. These methods produce interpretable layer extraction, but their dependence on handcrafted assumptions limits robustness and scalability to high-level semantic layers. Recent data-driven methods move toward modular parsing pipelines. DreamLayer~\cite{huang2025dreamlayer} builds synthetic layered data via matting, while MULAN~\cite{mulan} combines open-vocabulary detection~\cite{yao2023detclipv2}, segmentation~\cite{kirillov2023segment}, depth estimation~\cite{ranftl2020towards}, and ordering heuristics~\cite{lee2022instance} for layered decomposition in unconstrained scenes. Interactive matting approaches such as Click2Trimap~\cite{zhang2025trimapclick} further reduce the annotation effort by learning trimaps from minimal user clicks. Although effective in natural images~\cite{yang2024generative}, most of such methods focus on object-level foreground/background separation and cannot be directly transferred to graphic design, where typography, style effects, and structured layout are the core. Beyond decomposition, a complementary line of low-level image processing improves pixel-level quality through learnable image enhancement~\cite{liu2024pixel} and bit-depth enhancement~\cite{liu2026realworld, liu2025learning, liu2024multistage}; in contrast, our work targets semantic layer-level editability rather than appearance refinement.

Unlike natural images, in graphic design image decomposition, recent methods increasingly rely on MLLMs to first predict a draft protocol and then recover layers through VQGAN decoding or "first matting and then image inpainting" framework~\cite{niedecomposition, chen2025rethinking, van2017vqvae, esser2021vqgan, layerd}. These design is flexible but vulnerable to stage-wise error accumulation: protocol errors propagate to image matting, and image inpainting~\cite{zhang2024mutualinpaint} often introduces texture inconsistency or semantic drift. In addition, fused artistic typography and fine-grained editable attributes remain difficult to preserve in cascaded pipelines.

Our method addresses these gaps with a hybrid formulation: VLM-based text rendering protocol prediction for editable typography, and multi-branch diffusion generation for background/sticker layers with design-specific LoRA adaptation~\cite{hu2022lora, ding2023parameter}. The proposed method avoids a heavy dependence on cascaded post-processing while improving fidelity and editability in diverse design styles.

\begin{figure*}[!t]
    \centering
    \includegraphics[width=0.99\linewidth]{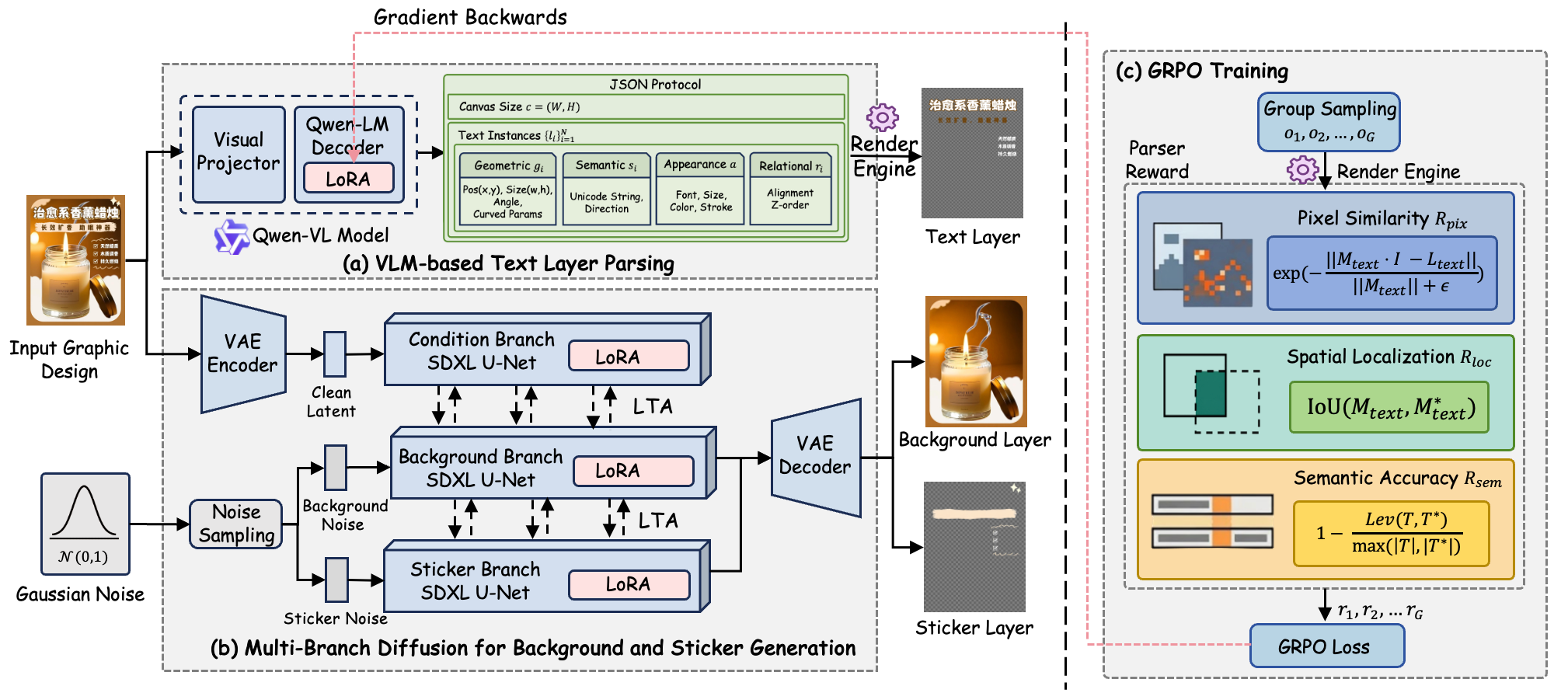}
    \caption{\textbf{Overview of the proposed CreatiParser framework.}
The framework comprises three components:
\textbf{(a)~VLM-based Text Layer Parsing} module (upper left), where a QwenLM-based multimodal decoder with LoRA generates text rendering protocols from the input graphic design, then with render engine to generate text layer;
\textbf{(b)~Multi-branch Diffusion} module (lower left), where three SDXL U-Net branches with Layer Token Attention (LTA) and LoRA adapters generate disentangled background and sticker layers;
\textbf{(c)~GRPO Training} (right), where the text parsing module is optimized via GRPO. Gradient backpropagation updates only the QwenLM LoRA weights while all other components remain frozen.}    \label{fig:framework}
\end{figure*}

\subsection{Reinforcement Learning on Visual Language Models}

Reinforcement learning has become an important post-training paradigm for visual language models, especially for reducing hallucination and improving response reliability. Early studies adapted RLHF to multimodal settings by introducing image-grounded factual rewards and correction-oriented preference supervision~\cite{sun2023farlhf, yu2024rlhfv}. Subsequent work explored AI-generated feedback as a scalable alternative to human-only annotation, showing that trustworthiness can be improved with open-source feedback pipelines~\cite{yu2025rlaifv}. Recent multimodal alignment frameworks further expand this line from safety-centric correction to broader capability alignment under unified preference optimization objectives~\cite{zhang2025mmrlhf}. These efforts establish RL-based post-training as a practical complement to supervised instruction tuning for VLM alignment.

In parallel, R1-style training has shifted attention toward rule-based and verifiable rewards for perception-intensive reasoning tasks. Perception-R1~\cite{yu2025perceptionr1} and related grounded reasoning studies~\cite{cao2026groundr1} show that rewards targeted at visual evidence usage can improve grounding quality beyond generic response-level preferences. For video understanding, VideoChat-R1~\cite{li2025videochatr1} and Video-R1~\cite{feng2025videor1} introduce reinforcement fine-tuning strategies that incorporate spatio-temporal constraints and temporal-aware policy updates. OpenThinkIMG~\cite{su2025openthinkimg} further explores tool-augmented RL for visual reasoning, emphasizing interaction-aware reward signals instead of static text-only feedback. Different from these general-purpose VLM alignment settings, our method applies GRPO to text rendering protocol prediction in graphic design parsing, where rewards are renderer-grounded and task-specific (pixel, layout, and semantic consistency) rather than open-ended conversational preference scores. More broadly, our setting connects to fine-grained multimodal understanding and localization, which has been extensively studied for video moment retrieval~\cite{liu2025aggregate}, joint moment retrieval and highlight detection~\cite{liu2025whatwhere}, referring video object segmentation~\cite{wang2025weakly}, text-queried action localization~\cite{su2023sequence}, visual object tracking~\cite{liu2024siamdmu}, emotional and subjective video captioning~\cite{ye2024dual, ye2025multi, chen2026subjective}, and video summarization~\cite{ye2025improving}. Reinforcement learning and preference modeling have likewise proven effective for interactive and cross-domain recommendation~\cite{nie2024knowledge, liu2024interintra, liu2024graph}, where aligning model behavior with user preferences is central---an objective we share but realize through renderer-grounded rewards for graphic design parsing.

\section{Method}

\subsection{Problem Formulation and Overview}

Given a rasterized graphic design image $I$, our goal is to decompose it into a set of editable and semantically disentangled layers to make designers flexibly adjust each component:
\begin{equation}
\mathcal{L} = \{L_{\text{text}}, L_{\text{sticker}}, L_{\text{background}}\},
\end{equation}
where $L_{\text{text}}$, $L_{\text{sticker}}$, $L_{\text{background}}$ correspond to the text layer, the sticker layer, and the background layer, respectively. 

Each layer is defined as follows:

\textbf{Text Layer} $L_{\text{text}}$: The editable layer that contains all typographic elements in the design. 

\textbf{Sticker Layer} $L_{\text{sticker}}$: An RGBA image representing all non-textual foreground elements, such as decorative lines, geometric shapes, icons, illustrative embellishments.

\textbf{Background Layer} $L_{\text{background}}$: A full-resolution RGB image that carries the global visual context of the design, including base colors, textures, and photographic imagery.

Graphic design parsing requires recovering both the editable text structure and the disentangled visual layers. To this end, we adopt a hybrid framework with three components: (1) A multi-branch diffusion model for background and sticker generation; (2) A vision--language text parser based on Qwen3-VL for structured text prediction; and (3) A reward-guided optimization scheme for training the text parser. An overview of the framework is illustrated in Fig.~\ref{fig:framework}.

During inference, the input image is fed into the diffusion branch and the VLM branch in parallel. The diffusion branch predicts the sticker and background layers, while the VLM branch outputs a text rendering protocol, which is converted by a render engine into the text layer. The three outputs together form the final layered parsing result.

\subsection{Multi-Branch Diffusion for Background and Sticker Generation}

As defined in Sec.~3.1, the background and sticker layers capture complementary non-textual content. The background layer provides a global visual canvas, while the sticker layer isolates localized decorative foreground elements. We model the generation of both layers using a diffusion-based generative framework. Formally, given an input design image $I$, our objective is to jointly generate $L_{\text{background}}$ (full-resolution RGB) and $L_{\text{sticker}}$ (RGBA with explicit alpha channel).

\subsubsection{Multi-Branch Diffusion Architecture}

Stable Diffusion has demonstrated strong generative priors for various image synthesis tasks~\cite{jiang2024animediff, xu2024sgdm}. Thus, we adopt Stable Diffusion XL~\cite{sdxl} as the backbone in this work, and construct a three-branch architecture to jointly model background and sticker generation. The three branches share the same U-Net topology but differ in their conditioning inputs:

\textbf{Condition branch}: Receives the latent representation $z^{(c)} = \mathcal{E}(I)$ of the clean input design image through the VAE encoder $\mathcal{E}$, where knowledge is then passed to the Background/Sticker branches.

\textbf{Background branch}: Receives noised latents $z^{(b)}_t$ corresponding to the background layer at diffusion timestep $t$.

\textbf{Sticker branch}: Receives noised latents $z^{(s)}_t$ corresponding to the sticker layer in RGBA space.

Each branch follows the standard SDXL U-Net structure with independent self-attention and cross-attention modules.

To adapt the diffusion model to the graphic design domain efficiently, each branch is equipped with an independent Low-Rank Adaptation (LoRA) module. Crucially, the three branches \emph{share a single frozen copy of the SDXL base U-Net weights} rather than instantiating three independent U-Net copies; branch specialization is achieved solely through the separate LoRA modules. LoRA is applied to all attention projection matrices ($W_Q, W_K, W_V, W_O$) in the U-Net. All LoRA parameters are optimized during training, while the original SDXL weights remain frozen. This weight-sharing design keeps the additional parameter count modest---only the LoRA adapters (and the LTA modules) are trainable and stored per branch, while the heavy backbone is shared---and enables each branch to specialize in its target layer without degrading the pretrained generative prior.

The sticker layer is generated as a transparent image with an explicit alpha channel. To support RGBA generation in latent space, following LayerDiffuse \cite{zhang2024transparent}, we extend the SDXL VAE architecture to handle 4-channel inputs and generate transparent. The diffusion process operates in the latent RGBA space, allowing the model to jointly synthesize color and transparency. This design enables faithful generation of decorative elements with soft boundaries and partial transparency, which are common in real-world graphic design assets.

\subsubsection{Layer Token Attention Mechanism}

\begin{figure}
    \centering
    \includegraphics[width=0.9\linewidth]{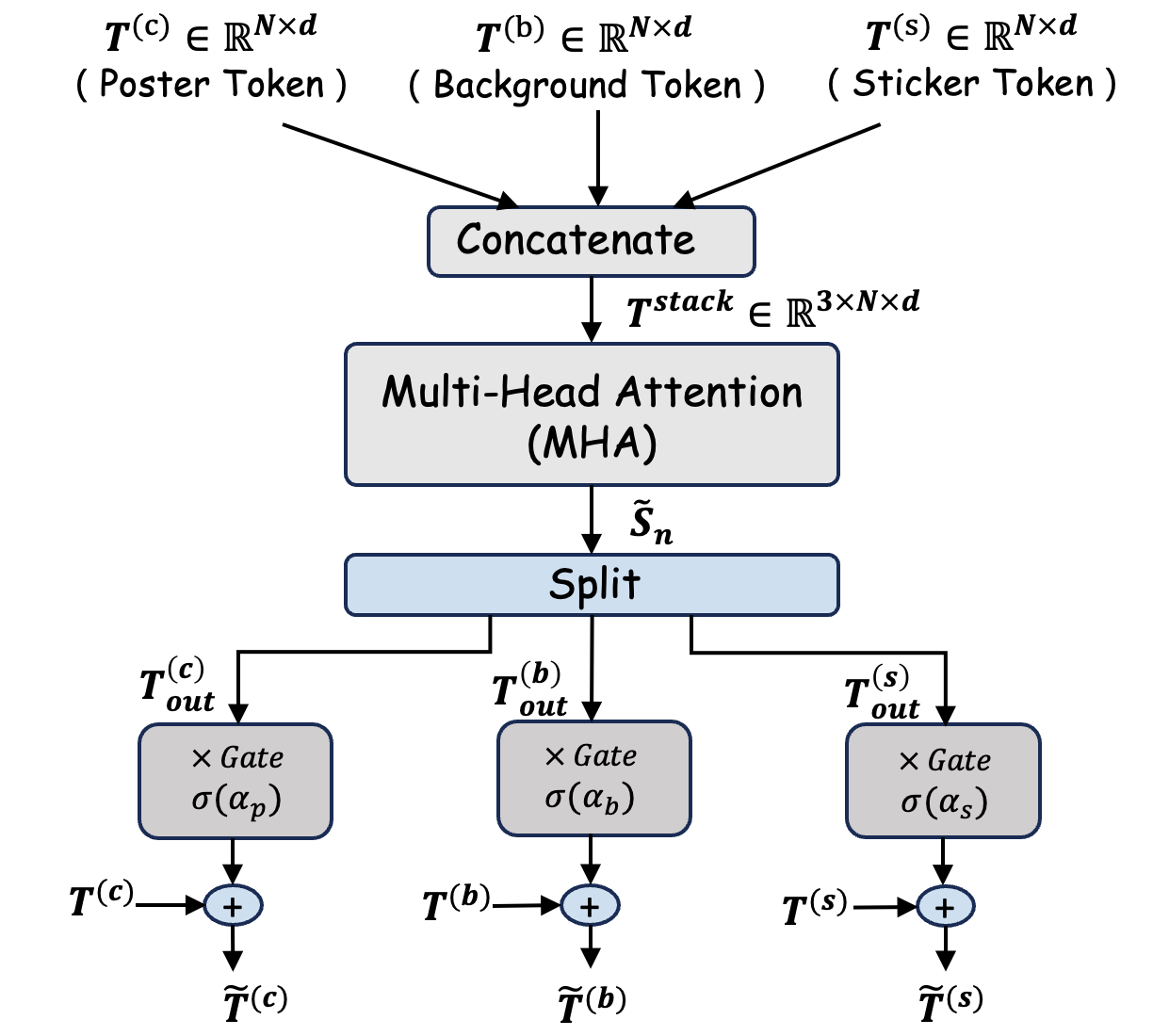}

    \caption{Illustration of Layer Token Attention (LTA). Tokens from the three branches are stacked into a $3 \times N \times d$ tensor, and self-attention is applied only along the branch dimension for each spatial location, enabling position-wise cross-branch information exchange.}
    \label{fig:LTA}
\end{figure}

To enable effective information exchange across branches while maintaining layer-specific generation capacity, we introduce Layer Token Attention (LTA) at every U-Net block, as illustrated in Fig.~\ref{fig:LTA}. Let $\mathbf{T}^{(c)} \in \mathbb{R}^{N_c \times d}$, $\mathbf{T}^{(b)} \in \mathbb{R}^{N_b \times d}$, and $\mathbf{T}^{(s)} \in \mathbb{R}^{N_s \times d}$ denote the token embeddings produced by the condition, background, and sticker branches after their respective self-attention computation, where $N_c, N_b, N_s$ are the token counts and $d$ is the feature dimension.

Since the three branches operate on the same spatial resolution, we have $N_c = N_b = N_s = N$. Instead of flattening all tokens jointly, LTA stacks branch features into a tensor
\begin{equation}
\mathbf{T}^{\text{stack}} = \text{Stack}(\mathbf{T}^{(c)}, \mathbf{T}^{(b)}, \mathbf{T}^{(s)}) \in \mathbb{R}^{3 \times N \times d},
\end{equation}
and applies self-attention along the branch dimension. For each spatial position $n \in \{1, \dots, N\}$, we form
\begin{equation}
\mathbf{S}_n = [\mathbf{T}^{(c)}_n; \mathbf{T}^{(b)}_n; \mathbf{T}^{(s)}_n] \in \mathbb{R}^{3 \times d},
\end{equation}
\begin{equation}
\hat{\mathbf{S}}_n = \text{MHA}(\mathbf{S}_n W_Q, \mathbf{S}_n W_K, \mathbf{S}_n W_V),
\end{equation}
where multi-head attention (MHA) is performed only among the three branch tokens at the same spatial location. Therefore, each token interacts only with its corresponding position in the other branches, rather than with all $3N$ tokens globally.

The fused outputs are redistributed back to the three branches. For $n = 1, \dots, N$, the formulation is:
\begin{equation}
\mathbf{T}^{(c)}_{\text{out}, n}, \mathbf{T}^{(b)}_{\text{out}, n}, \mathbf{T}^{(s)}_{\text{out}, n} = \text{Split}(\hat{\mathbf{S}}_n),
\end{equation}
These updated tokens are then combined with the original branch tokens via a learnable gating mechanism.
\begin{equation}
\tilde{\mathbf{T}}^{(k)} = \mathbf{T}^{(k)} + \sigma(\alpha_k) \cdot \mathbf{T}^{(k)}_{\text{out}}, \quad k \in \{c, b, s\},
\end{equation}
where $\alpha_k$ is a learnable scalar, allowing gradual integration of cross-branch information during training. This design enables fine-grained interaction among the three branches while preserving spatial correspondence.


\vspace{-10pt}

\subsection{VLM-based Text Layer Parsing via Text Rendering Protocols}
\label{sec:text_protocol}

Text elements in graphic design images exhibit strong structural regularities, including explicit geometry, typography, and rendering attributes. Instead of reconstructing text purely at the pixel level, we predict a text rendering protocol to preserve editability.

\subsubsection{Text Rendering Protocol}

Given an input image $I$, we represent the text layer by a protocol
\begin{equation}
P_{\text{text}} = \{c, \{\ell_i\}_{i=1}^{N}\},
\end{equation}
where $c = (W, H)$ denotes the canvas size, and each $\ell_i$ corresponds to a single text instance.

Each text instance $\ell_i$ is defined as
\begin{equation}
\ell_i = (g_i, s_i, a_i, r_i),
\end{equation}
where each element is a set that contains the following attributes:

\textbf{Geometric attributes} $g_i$: absolute position $(x, y)$, bounding box size $(w, h)$, orientation angle $\theta \in [0, 2\pi)$, and bending parameters $\mathbf{b}_i$. For curved or warped text commonly found in graphic designs, we parameterize the text path using a cubic B\'{e}zier curve. The bending parameters $\mathbf{b}_i$ are defined as:
\begin{equation}
\mathbf{b}_i = \{(p_0, p_1, p_2, p_3), \tau\},
\end{equation}
where $p_0, p_1, p_2, p_3 \in \mathbb{R}^2$ are the four control points of the B\'{e}zier curve, and $\tau \in \{0, 1\}$ indicates whether the text follows the curve ($\tau=1$) or remains straight ($\tau=0$). The curve is parameterized as:
\begin{equation}
B(t) = (1-t)^3 p_0 + 3(1-t)^2 t p_1 + 3(1-t)t^2 p_2 + t^3 p_3, \quad t \in [0, 1].
\end{equation}
For straight text, we set $\tau=0$ and the bending parameters are ignored during rendering

\textbf{Semantic attributes} $s_i$: Unicode text string and reading direction (left-to-right or right-to-left).

\textbf{Appearance attributes} $a_i$: font identifier $f_i$, font size, text color (RGB or gradient specification), stroke width and color, shadow configuration (color, offset angle, blur radius), line height, character spacing, and emphasis styles (italic, bold, underline).

\textbf{Relational attributes} $r_i$: alignment mode (left, center, right, justify) and z-order for layer stacking.

\subsubsection{Vision--Language Protocol Prediction}

We employ a pretrained vision--language model, Qwen3-VL~\cite{yang2025qwen3}, to infer the text rendering protocol directly from the input image:
\begin{equation}
P_{\text{text}} = F_{\text{VL}}(I).
\end{equation}

The model jointly reasons over visual layout and textual semantics, allowing it to recover fine-grained attributes such as font style, alignment, and decorative effects. We fine-tune Qwen3-VL by LoRA~\cite{hu2022lora} on our self-collected Parser-40K dataset with paired (image, protocol) annotations. The output protocol follows a structured JSON schema. Then we use a render engine to convert the protocol to the text layer.

\subsection{ParserReward-Guided GRPO for Text Parsing}

To improve text accuracy and editability, we optimize the Qwen3-VL text parser with Group Relative Policy Optimization (GRPO). The reward evaluates whether the predicted protocol, after deterministic rendering, preserves the original text appearance, layout, and semantics.

\subsubsection{Parser Reward}

Given the rendered text layer $L_{\text{text}}$, we define three complementary reward terms:
\begin{equation}
R_{\text{pix}} = \exp\left(-\frac{\|M_{\text{text}} \odot I - L_{\text{text}}\|_1}{\|M_{\text{text}}\|_1 + \epsilon}\right),
\end{equation}
\begin{equation}
R_{\text{loc}} = \text{IoU}(M_{\text{text}}, M^*_{\text{text}}),
\end{equation}
\begin{equation}
R_{\text{sem}} = \text{Sim}_{\text{edit}}(T_{\text{protocol}}, T^*),
\end{equation}
where $M_{\text{text}}$ and $M^*_{\text{text}}$ are the predicted and reference text masks, $T_{\text{protocol}}$ and $T^*$ are the predicted and reference text strings, and $\epsilon=10^{-6}$ is a small constant added for numerical stability. We use normalized Levenshtein similarity
\begin{equation}
\text{Sim}_{\text{edit}}(a, b) = 1 - \frac{\text{Lev}(a, b)}{\max(|a|, |b|)},
\end{equation}
and combine the three terms into a weight-normalized weighted average to obtain the total Reward $R_{\text{text}}$.
\begin{equation}
R_{\text{text}} = \frac{\lambda_{\text{pix}} R_{\text{pix}} + \lambda_{\text{loc}} R_{\text{loc}} + \lambda_{\text{sem}} R_{\text{sem}}}{\lambda_{\text{pix}} + \lambda_{\text{loc}} + \lambda_{\text{sem}}},
\end{equation}
where $\lambda_{\text{pix}}$, $\lambda_{\text{loc}}$, and $\lambda_{\text{sem}}$ balance the three reward terms. Normalizing by the sum of the weights makes $R_{\text{text}}$ a convex combination of the three component rewards, so that $R_{\text{text}}\in[0,1]$ regardless of the weight scale (since each term lies in $[0,1]$) and the reward magnitude stays comparable across different weightings. We set $\lambda_{\text{pix}}=\lambda_{\text{loc}}=\lambda_{\text{sem}}=1$ by default, under which $R_{\text{text}}$ reduces to the simple average of the three terms; a sensitivity analysis over alternative weightings is provided in Sec.~\ref{sec:ablation} (Table~\ref{tab:ablation_reward}), which confirms that the equal-weight setting yields the best overall balance.

\begin{algorithm}[t]
\caption{ParserReward-guided GRPO for Qwen3-VL text parsing}
\label{alg:grpo}
\begin{algorithmic}[1]
\REQUIRE Design images $\mathcal{D}$; group size $K=16$; learning rate $\eta=10^{-4}$; total steps $T=2000$
\STATE Initialize LoRA adapter parameters $\theta$; freeze backbone $\phi$
\STATE Initialize reference policy $\pi_{\text{ref}} \leftarrow \pi_\theta$
\FOR{$t = 1$ to $T$}
    \STATE Sample mini-batch $\{I_i\}_{i=1}^{B} \sim \mathcal{D}$ with $B=32$
    \FOR{each design image $I_i$}
        \STATE Sample $K$ protocols $\{P_i^{(k)}\}_{k=1}^{K}$ from $\pi_\theta(\cdot|I_i)$
        \FOR{$k = 1$ to $K$}
            \STATE Render text layer: $L_{\text{text}}^{(k)} \leftarrow \text{Render}(P_i^{(k)})$
            \STATE Compute reward: $r_i^{(k)} \leftarrow R_{\text{text}}(I_i, L_{\text{text}}^{(k)})$
        \ENDFOR
        \STATE Compute advantages $\{A_i^{(k)}\}$ via group normalization
    \ENDFOR
    \STATE Compute $\mathcal{L}_{\text{GRPO}}$ with clipped ratios
    \STATE Add KL regularization: $\mathcal{L}_{\text{total}} \leftarrow \mathcal{L}_{\text{GRPO}} + \beta \cdot D_{\text{KL}}$
    \STATE Update: $\theta \leftarrow \theta - \eta \nabla_\theta \mathcal{L}_{\text{total}}$
\ENDFOR
\RETURN Optimized adapter parameters $\theta$
\end{algorithmic}
\end{algorithm}

\subsubsection{GRPO Optimization}

Given an input design image $I$, we sample $K$ candidate text rendering protocols $\{P^{(k)}\}_{k=1}^{K}$ from Qwen3-VL using temperature sampling ($\tau = 0.8$). Each protocol is rendered into a text layer and scored by
\begin{equation}
r^{(k)} = R_{\text{text}}(I, \text{Render}(P^{(k)})).
\end{equation}

We then normalize rewards within each group:
\begin{equation}
\label{eq:advantage}
A^{(k)} = \frac{r^{(k)} - \mu_r}{\sigma_r + \epsilon},
\end{equation}
where $\mu_r = \frac{1}{K}\sum_{k=1}^{K} r^{(k)}$ and $\sigma_r = \sqrt{\frac{1}{K}\sum_{k=1}^{K}(r^{(k)} - \mu_r)^2}$. These within-group statistics $(\mu_r, \sigma_r)$---computed over the $K$ candidate protocols sampled for the \emph{same} design image---are exactly the quantities used to compute the advantages $A^{(k)}$ in line~11 of Algorithm~\ref{alg:grpo}; normalizing within each group rather than across images keeps the advantage scale comparable across designs of differing difficulty. The policy $\pi_\theta$ (Qwen3-VL with LoRA adapters) is updated using
\begin{equation}
\mathcal{L}_{\text{G}} = -\mathbb{E}\sum_{k=1}^{K} \min\left(\rho^{(k)} A^{(k)}, \text{clip}(\rho^{(k)}, 1-\epsilon_c, 1+\epsilon_c) A^{(k)}\right),
\end{equation}
where $\rho^{(k)} = \frac{\pi_\theta(P^{(k)}|I)}{\pi_{\theta_{\text{old}}}(P^{(k)}|I)}$ is the probability ratio, and $\epsilon_c = 0.2$ is the clipping threshold.

To limit drift from the supervised initialization, we add a KL regularizer:
\begin{equation}
\mathcal{L}_{\text{total}} = \mathcal{L}_{\text{G}} + \beta \cdot D_{\text{KL}}(\pi_\theta \| \pi_{\text{ref}}),
\end{equation}
where $\pi_{\text{ref}}$ is the reference policy (initial fine-tuned model) and $\beta = 0.01$ controls the regularization strength.

\section{Experiments}

In this section, we conduct extensive experiments to evaluate the effectiveness, robustness, and generalization capability of the proposed framework for graphic design parsing. The experiments are designed to assess (i) layer decomposition accuracy, (ii) text editability and rendering fidelity, (iii) background and sticker generation quality, and (iv) cross-dataset generalization.

\begin{figure*}
    \centering
    \includegraphics[width=0.85\linewidth]{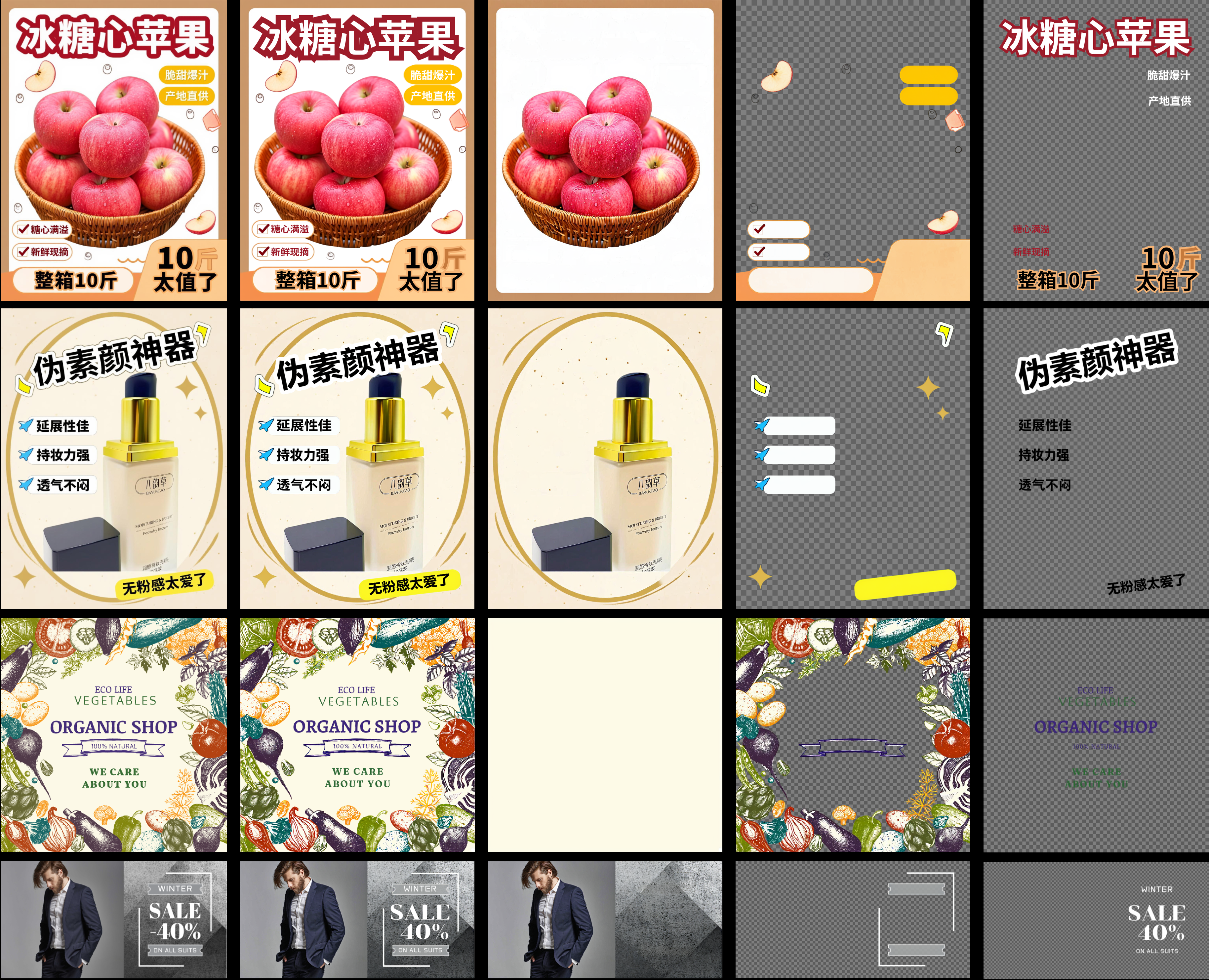}
    \caption{From left to right column, we show: (a) the input poster image; (b) the reconstructed design by compositing the parsed layers; (c) the extracted background; (d) the extracted sticker layer; and (e) the extracted text layer. Our method accurately separates semantic layers, including text, decorative sticker elements, and background regions across diverse design styles, enabling clean layer extraction and faithful reconstruction.}
    \label{fig:visualize}
\end{figure*}

\subsection{Datasets}
\label{sec:datasets}

We conduct experiments on two datasets covering both professionally designed graphics and AI-generated raster designs.

\subsubsection{Parser-40K Dataset}

We construct the Parser-40K dataset, a large-scale graphic design dataset specifically curated for this work. The images are collected from professional design platforms and licensed stock design libraries, with explicit permission for research use, and each design is exported in PSD format, providing native layer decomposition. Since the source files already contain editable text objects and a complete layer structure, no additional manual annotation is required. We directly parse the layered design files and convert them into our unified three-layer representation: all layers of type \texttt{sticker} within a design are composited into a single sticker layer, while the original text metadata are converted into our text rendering protocol, and the remaining visual content forms the background supervision.

\textbf{Dataset Statistics.} Parser-40K contains a total of $41{,}317$ layered designs, split into $37{,}185$ training samples, $2{,}066$ validation samples, and $2{,}066$ test samples (90\%/5\%/5\%) with no design overlap across splits. The canvas resolutions span common social-media and print aspect ratios, ranging from $512{\times}512$ to $2048{\times}2048$. Each design contains on average $4.7$ text instances ($1$--$23$ per design), $6.2$ sticker elements, and $1$ background layer. Aggregated across the corpus, the dataset comprises $194{,}203$ text instances and $256{,}149$ sticker elements. The text instances cover $612$ distinct font families, with a long-tailed frequency distribution typical of real-world design assets. Regarding text geometry, $83.6\%$ of text instances are straight ($\tau{=}0$) and $16.4\%$ are curved/warped ($\tau{=}1$), reflecting the prevalence of decorative arc- and path-following typography in graphic design. By design category, the corpus spans posters/flyers (38\%), social-media creatives (29\%), banners/ads (21\%), and cards/invitations (12\%). Detailed per-attribute distributions are provided in the supplementary material.

\subsubsection{Crello Dataset}

The Crello~\cite{crello} dataset is a large-scale real-world design platform dataset featuring highly diverse templates, decorative elements, artistic typography, and complex visual compositions. Importantly, Crello data is not used during training. We directly apply our trained model to Crello in a zero-shot manner to evaluate its generalization ability across unseen styles and layouts.

\subsection{Evaluation Metrics}

We evaluate graphic design parsing using comprehensive metrics covering reconstruction accuracy, text editability, and perceptual quality.

\textbf{Layer Reconstruction Accuracy}: We compute IoU between predicted and ground-truth masks for text (T-IoU) and sticker (S-IoU) layers.

\textbf{Text Editability}: Since our method predicts a structured text rendering protocol (Sec.~\ref{sec:text_protocol}), we directly evaluate the accuracy of the predicted protocol fields against ground-truth annotations.
\textit{Font Accuracy} (Font) is reported separately, as font identification constitutes a high-cardinality discrete classification over hundreds of font families and is the primary determinant of text visual identity; it is evaluated by exact match between the predicted font identifier $f_i$ and the ground-truth label.
\textit{Attribute Accuracy} (Attr.) reports the mean per-field accuracy over the remaining appearance attributes $a_i$ and relational attributes $r_i$ of the protocol.
Discrete fields---emphasis styles (italic, bold, underline) and alignment mode (left, center, right, justify)---are evaluated by exact match.
Continuous fields---font size, text color (RGB), stroke width and color, shadow configuration (offset angle, blur radius, color), line height, and character spacing---are judged correct if the predicted value falls within a threshold.

\textbf{Pixel-Level Reconstruction}: We report RGB L1 error between the generated and ground-truth layers, computed separately for the text, sticker, and background layers, as well as the average across all layers.

\textbf{Perceptual Quality}: To complement the pixel-level L1 error, which is a limited discriminator of visual quality and can miss boundary artifacts such as alpha bleeding and color fringing, we additionally report a comprehensive suite of perceptual metrics in Table~\ref{tab:perceptual}: LPIPS~\cite{zhang2018lpips}, SSIM~\cite{wang2004ssim}, and PSNR computed per layer and over the full composited design, together with FID~\cite{heusel2017fid} over composited designs. For the RGBA sticker layer we evaluate both the composited RGB render (under the standard alpha-over operation against a neutral background) and the alpha channel itself; the alpha matte is assessed separately in Table~\ref{tab:alpha} via alpha SSIM, PSNR, MAE, and alpha-IoU.

\textbf{Evaluation Protocol}: Unless otherwise stated, all reported metrics are computed over the \emph{full} test set ($2{,}066$ Parser-40K test designs and the complete Crello evaluation split), not a sampled subset. To assess statistical reliability, every CreatiParser variant is trained and evaluated with three random seeds. The main tables report the mean over the three runs; the corresponding standard deviations are summarized in Table~\ref{tab:seeds} and are small relative to the gaps between methods, confirming that the reported improvements are stable across seeds.

\subsection{Implementation Details}
CreatiParser consists of two main components: a Qwen3-VL-based text parsing module and a
diffusion-based background and sticker generation module.

The text parsing module is fine-tuned from Qwen3-VL-8b with LoRA using paired design images and
text rendering protocols. Training is performed with cross-entropy loss on both textual
content and structural attributes.

For background and sticker generation, we employ a three-branch SDXL backbone.
Each branch is equipped with independent LoRA adapters to inject domain-specific
knowledge while preserving pretrained priors. 

ParserReward-guided optimization is implemented using GRPO. For each training image, multiple candidate text rendering protocols are sampled from Qwen3-VL and evaluated by ParserReward, and the VLM policy is then refined through group-relative advantage estimation with clipped ratio updates.

\textbf{Hardware and Training Configuration.} All experiments are conducted on a server with $8\times$ NVIDIA A100 80GB GPUs. The SFT stage of Qwen3-VL-8B (LoRA rank 256) is trained for $3$ epochs on Parser-40K. The GRPO stage uses group size $K{=}16$, batch size $B{=}32$ design images per step, and $T{=}2000$ steps, with $K{=}16$ protocol rollouts rendered and scored per image per step; this stage runs on the $8\times$A100 node and completes in approximately $320$ GPU-hours. The three-branch SDXL diffusion module (rank 64 per branch) is trained separately. To ensure reproducibility, all reported results are averaged over three random seeds $\{0, 1, 2\}$.

\begin{table*}[t]
\centering
\caption{Comprehensive quantitative comparison on layer reconstruction, text editability, and perceptual quality. Results are reported on Parser-40K testset and Crello dataset (zero-shot). $\uparrow$: higher is better; $\downarrow$: lower is better. Best results in \textbf{bold}.}
\label{tab:main_results}
\resizebox{0.85\linewidth}{!}
{
\begin{tabular}{ll|cc|cc|cccc}
\toprule
\multirow{2}{*}{Dataset} & \multirow{2}{*}{Method} & \multicolumn{2}{c|}{Layer IoU $\uparrow$} & \multicolumn{2}{c|}{Text Editability $\uparrow$} & \multicolumn{4}{c}{RGB L1 $\downarrow$}\\
& & T-IoU & S-IoU & Font & Attr & Text & Sticker & BG & Avg  \\
\midrule
\multirow{5}{*}{Parser-40K}
& Baseline & 0.712 & 0.438 & -- & -- & 0.0813 & 0.1034 & 0.0762 & 0.0870  \\
& LayerD & 0.794 & 0.743 & -- & -- & \textbf{0.0472} & 0.0396 & 0.0615 & 0.0494  \\
& CreatiParser & 0.873 & \textbf{0.862} & 0.824 & 0.891 & 0.0641 & \textbf{0.0271} & \textbf{0.0385} & 0.0435  \\
& CreatiParser-RL & \textbf{0.896} & \textbf{0.862} & \textbf{0.873} & \textbf{0.912} & 0.0567 & \textbf{0.0271} & \textbf{0.0385} & \textbf{0.0410} \\
\midrule
\multirow{5}{*}{Crello}
& Baseline & 0.741 & 0.463 & -- & -- & 0.0736 & 0.0948 & 0.0685 & 0.0790 \\
& LayerD & 0.822 & 0.764 & -- & -- & \textbf{0.0414} & 0.0337 & 0.0573 & 0.0441 \\
& CreatiParser & 0.892 & \textbf{0.871} & 0.812 & 0.883 & 0.0635 & \textbf{0.0243} & \textbf{0.0364} & 0.0415  \\
& CreatiParser-RL & \textbf{0.914} & \textbf{0.871} & \textbf{0.856} & \textbf{0.907} & 0.0618 & \textbf{0.0243} & \textbf{0.0364} & \textbf{0.0409} \\
\bottomrule
\end{tabular}}
\end{table*}

\begin{figure}[!t]
    \centering
    \includegraphics[width=0.8\linewidth]{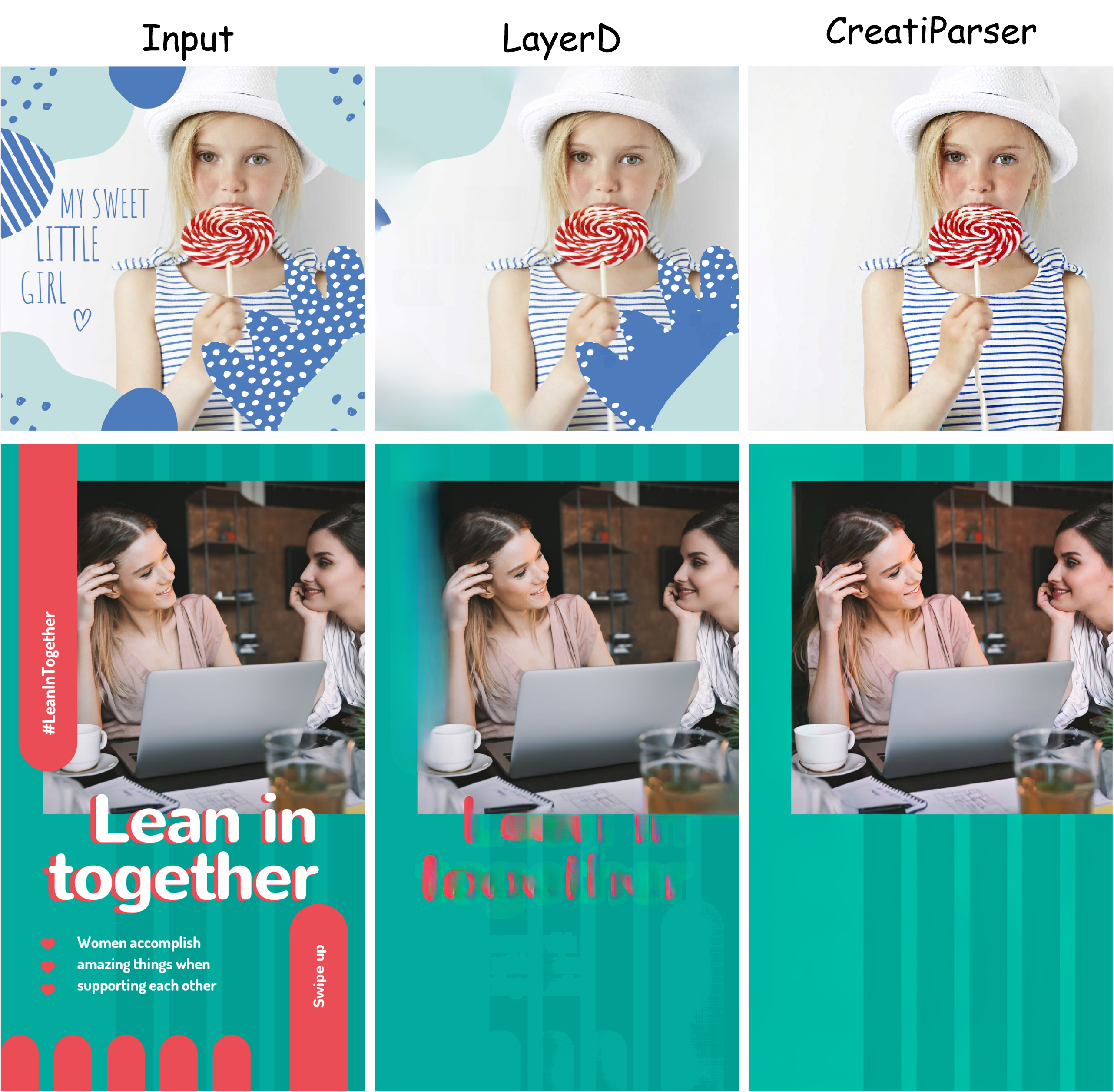}
    \caption{Comparison of background layer generation quality by CreatiParser and LayerD. The leftmost images are the input image. The 2nd and 3rd columns are the decomposition results of the background layer.}
    \label{fig:background_comp}
\end{figure}

\subsection{Comparison Methods}

We compare CreatiParser with baselines covering both traditional pipelines and recent generative approaches.

\textbf{Baseline}: A conventional multi-stage pipeline combining semantic segmentation with image inpainting. We leverage the Grounded-SAM-2~\cite{ren2024grounded} for text and sticker segmentation, and the LaMa~\cite{suvorov2022resolution} for background inpainting. All models leverage the official pre-trained weights.

\textbf{LayerD}~\cite{layerd}: A matting-first approach followed by background completion. We use the official code and pretrained weights released by the authors.

\textbf{Layer Category Normalization}: The compared methods, \emph{i.e.}, Baseline and LayerD, and the dataset, \emph{i.e.,} Crello, employ different layer taxonomies that cannot directly align with our three-category schema (text/sticker/background), thus necessitating a systematic normalization procedure for fair comparison. We leverage GPT-4V~\cite{openai2023gpt4v} as an automatic layer classifier by providing it with each layer image, along with its spatial context within the full design, and a structured prompt that specifies our category definitions. To validate the reliability of this automatic classification, we randomly sampled 200 layers, \emph{i.e.,} 100 from baseline outputs and 100 from the Crello dataset, and obtained the ground-truth labels through independent annotation by two expert annotators with design backgrounds, achieving an inter-annotator agreement of Cohen's $\kappa = 0.92$. A comparison of GPT-4V predictions against these consensus labels yields an overall classification accuracy of 94.5\%, with per-category precision and recall of 96.2\%/95.8\% for text, 91.3\%/90.7\% for stickers, and 95.1\%/96.4\% for background layers, respectively. Error analysis reveals that misclassifications primarily occur for heavily stylized decorative text elements that exhibit visual characteristics similar to those of stickers. Importantly, this normalization procedure is applied uniformly across all baseline methods and Crello samples, ensuring that any residual classification errors do not systematically bias the comparative evaluation.

\subsection{Qualitative Results}
\label{sec:vis_comp}

\begin{figure}[!t]
    \centering
    \includegraphics[width=0.78\linewidth]{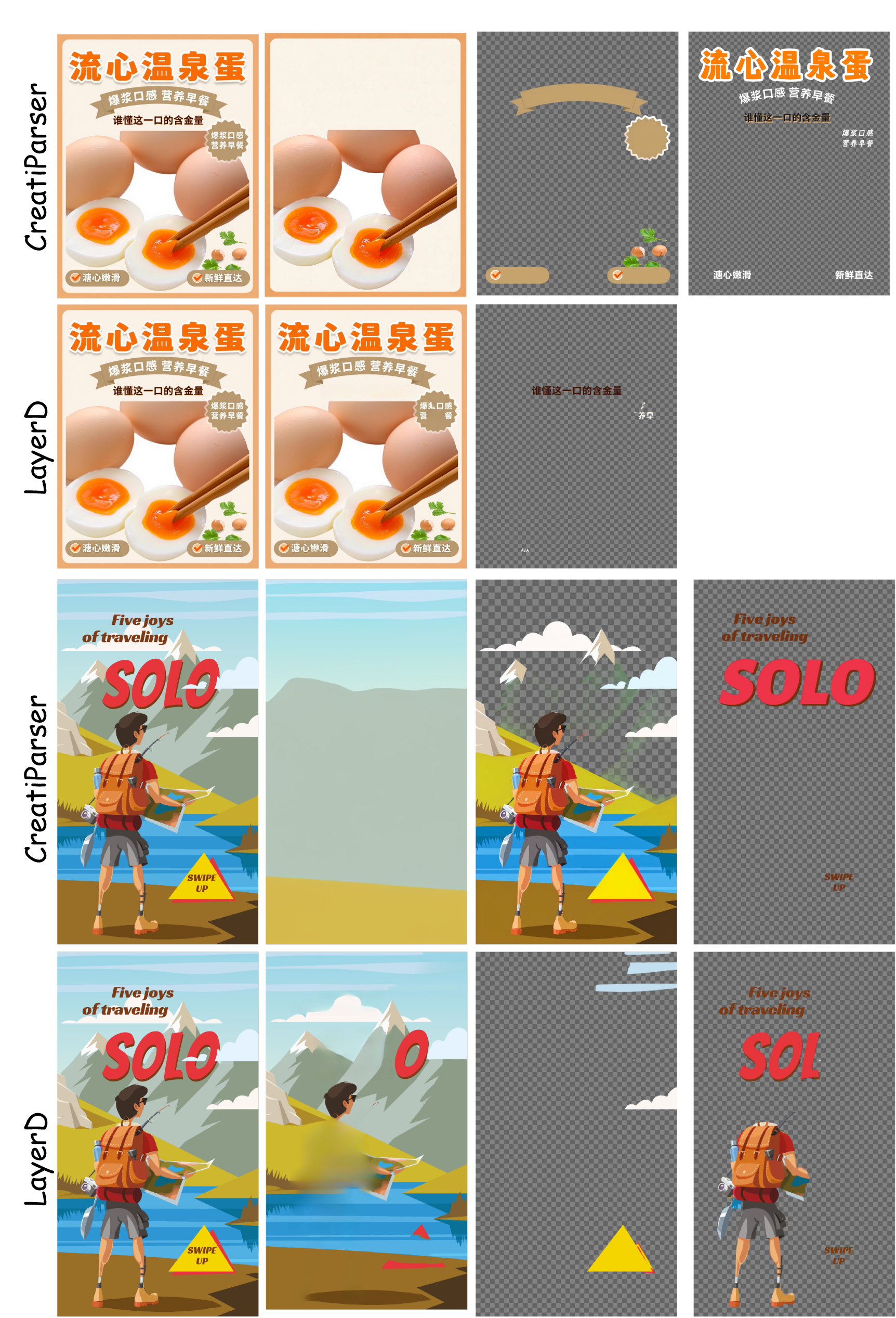}
    \caption{Visualization Comparison of full decomposition results between creatiparser and LayerD.}
    \label{fig:comparison_layerD}
\end{figure}

Fig.~\ref{fig:visualize} visualizes our semantic-layer decomposition across diverse poster styles. 
From left to right, we show: (a) input, (b) reconstruction by recompositing parsed layers, (c) background, (d) sticker, and (e) text. 
Our method yields clean, well-separated layers: backgrounds retain global templates and color tones, stickers preserve sharp shapes, and text regions remain compact and legible. 
The close match between the reconstruction and the input indicates faithful decomposition. Fig.~\ref{fig:background_comp} and Fig.~\ref{fig:comparison_layerD} further demonstrate robustness on photography-centric and illustration-based designs compared to the LayerD. Our results reduce common artifacts such as text leakage into backgrounds and confusion between stickers and typography, enabling reliable downstream editing (e.g., text replacement, sticker removal, and background re-layout).

\begin{table*}[t]
\centering

\caption{Perceptual-quality comparison on Parser-40K (test) and Crello (zero-shot). We report LPIPS, SSIM, and PSNR per layer (text, sticker RGB composite, background) and for the full composited design, plus FID over composited designs. LPIPS$\downarrow$ and FID$\downarrow$: lower is better; SSIM$\uparrow$ and PSNR$\uparrow$: higher is better. Best results in \textbf{bold}, second-best \underline{underlined}.\protect\footnotemark[2]}
\label{tab:perceptual}
\resizebox{\linewidth}{!}
{
\begin{tabular}{ll|ccc|ccc|ccc|cccc}
\toprule
\multirow{2}{*}{Dataset} & \multirow{2}{*}{Method} & \multicolumn{3}{c|}{Text} & \multicolumn{3}{c|}{Sticker (RGB)} & \multicolumn{3}{c|}{Background} & \multicolumn{4}{c}{Overall (composited)} \\
& & LPIPS$\downarrow$ & SSIM$\uparrow$ & PSNR$\uparrow$ & LPIPS$\downarrow$ & SSIM$\uparrow$ & PSNR$\uparrow$ & LPIPS$\downarrow$ & SSIM$\uparrow$ & PSNR$\uparrow$ & LPIPS$\downarrow$ & SSIM$\uparrow$ & PSNR$\uparrow$ & FID$\downarrow$ \\
\midrule
\multirow{4}{*}{Parser-40K}
& Baseline        & 0.142 & 0.842 & 22.4 & 0.198 & 0.781 & 19.8 & 0.176 & 0.803 & 21.2 & 0.169 & 0.812 & 21.6 & 38.4 \\
& LayerD          & \textbf{0.051} & \textbf{0.961} & \textbf{31.8} & \underline{0.087} & \underline{0.902} & \underline{26.3} & \underline{0.132} & \underline{0.871} & \underline{24.6} & 0.094 & 0.908 & 27.1 & 19.7 \\
& CreatiParser    & 0.068 & 0.944 & 29.6 & \textbf{0.054} & \textbf{0.948} & \textbf{30.1} & \textbf{0.071} & \textbf{0.931} & \textbf{28.9} & \underline{0.058} & \underline{0.951} & \underline{31.2} & \underline{11.2} \\
& CreatiParser-RL & \underline{0.061} & \underline{0.951} & \underline{30.5} & \textbf{0.054} & \textbf{0.948} & \textbf{30.1} & \textbf{0.071} & \textbf{0.931} & \textbf{28.9} & \textbf{0.052} & \textbf{0.957} & \textbf{31.8} & \textbf{9.8} \\
\midrule
\multirow{4}{*}{Crello}
& Baseline        & 0.135 & 0.851 & 22.9 & 0.189 & 0.792 & 20.3 & 0.168 & 0.812 & 21.7 & 0.161 & 0.821 & 22.1 & 35.1 \\
& LayerD          & \textbf{0.047} & \textbf{0.965} & \textbf{32.4} & \underline{0.081} & \underline{0.911} & \underline{26.9} & \underline{0.124} & \underline{0.878} & \underline{25.1} & 0.088 & 0.914 & 27.6 & 17.9 \\
& CreatiParser    & 0.064 & 0.948 & 30.1 & \textbf{0.049} & \textbf{0.953} & \textbf{30.6} & \textbf{0.066} & \textbf{0.936} & \textbf{29.4} & \underline{0.053} & \underline{0.956} & \underline{31.7} & \underline{9.8} \\
& CreatiParser-RL & \underline{0.058} & \underline{0.955} & \underline{31.0} & \textbf{0.049} & \textbf{0.953} & \textbf{30.6} & \textbf{0.066} & \textbf{0.936} & \textbf{29.4} & \textbf{0.048} & \textbf{0.961} & \textbf{32.3} & \textbf{8.6} \\
\bottomrule
\end{tabular}}
\end{table*}
\begin{table}[t]
\centering
\caption{Sticker alpha-channel evaluation on Parser-40K (test) and Crello (zero-shot). Alpha is treated as a single-channel matte. We report alpha SSIM, PSNR, mean absolute error (MAE), and the alpha-IoU at threshold $0.5$. CreatiParser$^{\ast}$ denotes both CreatiParser and CreatiParser-RL. Best in \textbf{bold}.}
\label{tab:alpha}
\resizebox{\columnwidth}{!}
{
\begin{tabular}{ll|cccc}
\toprule
Dataset & Method & $\alpha$-SSIM$\uparrow$ & $\alpha$-PSNR$\uparrow$ & $\alpha$-MAE$\downarrow$ & $\alpha$-IoU$\uparrow$ \\
\midrule
\multirow{2}{*}{Parser-40K}
& LayerD          & 0.872 & 24.1 & 0.0412 & 0.781 \\
& CreatiParser$^{\ast}$ & \textbf{0.931} & \textbf{28.7} & \textbf{0.0231} & \textbf{0.857} \\
\midrule
\multirow{2}{*}{Crello}
& LayerD          & 0.881 & 24.6 & 0.0398 & 0.792 \\
& CreatiParser$^{\ast}$ & \textbf{0.936} & \textbf{29.1} & \textbf{0.0219} & \textbf{0.871} \\
\bottomrule
\end{tabular}}
\end{table}

\subsection{Quantitative Results}

Table~\ref{tab:main_results} reports comprehensive quantitative comparisons. Overall, CreatiParser consistently outperforms all baseline methods across all metrics and datasets.

\textbf{Layer Reconstruction}: CreatiParser improves both text and sticker IoU over all baselines. On Parser-40K, CreatiParser attains 0.862 S-IoU and CreatiParser-RL further raises T-IoU to 0.896; the large sticker-IoU margin over the baselines ($0.438$/$0.743{\to}0.862$) reflects the benefit of RGBA-aware diffusion with explicit transparency modeling. We emphasize that, because GRPO updates only the text parser while the multi-branch diffusion model is kept frozen (Sec.~III-D, Algorithm~\ref{alg:grpo}), the sticker and background layers produced by CreatiParser and CreatiParser-RL are identical; consequently the diffusion-only metrics---S-IoU, Sticker/BG L1, and all alpha-matte scores (Table~\ref{tab:alpha})---are unchanged by RL, and CreatiParser-RL differs from CreatiParser only on the text-dependent metrics (T-IoU, Font, Attr., Text L1) and on the composited metrics that aggregate the refined text layer (Avg L1, the Overall columns, and FID in Table~\ref{tab:perceptual}). RGB L1 results in Table~\ref{tab:main_results} show the same trend. However, LayerD remains slightly better on text-layer L1, which we attribute to its use of a leading text segmentation model Hi-SAM for pixel-level extraction\cite{layerd}. In contrast, our method predicts a structured text protocol, enabling editable font and attribute control beyond pixel-level decomposition.

\footnotetext[2]{CreatiParser and CreatiParser-RL share the same diffusion model. Their Sticker and Background scores are identical.}

\textbf{Perceptual Quality}: Beyond pixel-level error, Table~\ref{tab:perceptual} and Table~\ref{tab:alpha} assess perceptual fidelity. In Table~\ref{tab:perceptual}, CreatiParser-RL attains the best overall LPIPS, SSIM, and PSNR and the lowest FID on both datasets (overall FID $9.8$ vs.\ LayerD's $19.7$ on Parser-40K, and $8.6$ vs.\ $17.9$ on Crello), with clear gains on the sticker and background layers. Consistent with the text-layer L1 result, LayerD retains a slight edge on the \emph{text} layer (text LPIPS $0.051$ vs.\ our $0.061$ on Parser-40K) owing to its dedicated pixel-level segmentation; our protocol-based prediction trades a small pixel-alignment gap for full editability. Table~\ref{tab:alpha} further shows that our generative RGBA design yields cleaner sticker boundaries: CreatiParser reaches $\alpha$-IoU $0.857$ and $\alpha$-MAE $0.0231$ on Parser-40K versus LayerD's $0.781$ and $0.0412$, directly reflecting the reduced alpha bleeding that RGB L1 alone cannot capture.

\begin{table}[!t]
\centering
\caption{Standard deviation of key metrics across three random seeds $\{0,1,2\}$ for CreatiParser-RL on Parser-40K. Standard deviations are small relative to the inter-method gaps in Table~\ref{tab:main_results}.}
\label{tab:seeds}
\resizebox{0.9\columnwidth}{!}
{
\begin{tabular}{lcccc}
\toprule
Metric & T-IoU & S-IoU & Font & RGB L1 (Avg) \\
\midrule
Mean        & 0.896 & 0.862 & 0.873 & 0.0410 \\
Std ($\pm$) & 0.004 & 0.005 & 0.006 & 0.0008 \\
\bottomrule
\end{tabular}}
\end{table}

\textbf{Text Editability}: Our VLM-based text parsing significantly outperforms image-based extraction methods. CreatiParser provides font identification (87.3\%) and style attribute prediction (91.2\%) that other methods cannot support. The GRPO optimization further improves all text metrics by 0.8--4.9\%.

\begin{figure}[!t]
    \centering
    \includegraphics[width=0.95\linewidth]{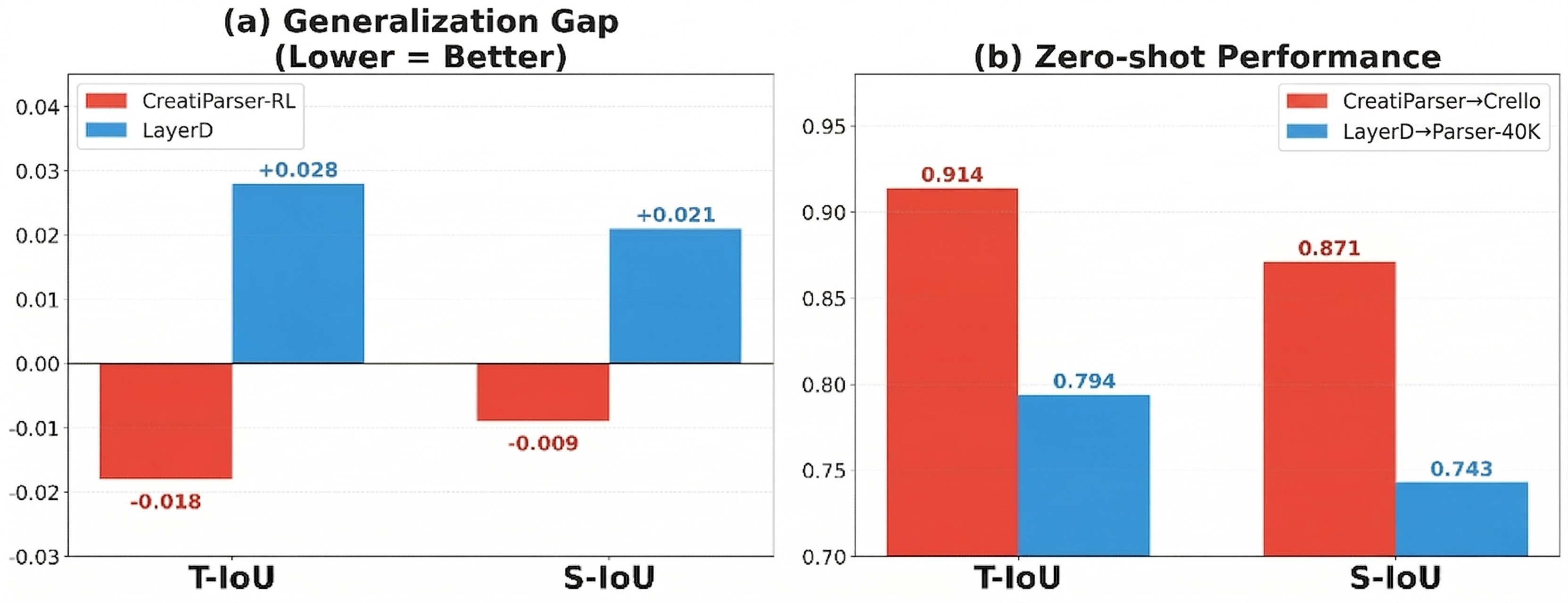}
    \caption{Generalization comparison between \textbf{CreatiParser} and \textbf{LayerD}.}
    \label{fig:generalize}
    \vspace{-10pt}
\end{figure}

\textbf{Zero-shot Generalization}: On the Crello dataset, which is not seen during training, the proposed method CreatiParser maintains or even exceeds its Parser-40K performance, e.g., T-IoU $0.892$ vs.\ $0.873$. As illustrated in Fig \ref{fig:generalize}, we draw the two metrics between CreatiParser and LayerD. The value in the left figure is defined as \textbf{seen metric - unseen metric}, where the smaller the absolute value of the difference, the more stable the model's generalization performance. We attribute this to the fact that the Crello dataset templates tend to have cleaner layouts and more standardized typography compared to the diverse and often noisy designs in Parser-40K, making them inherently easier to parse. This result demonstrates the strong generalization capability of the proposed framework CreatiParser to unseen design styles and layouts.

\subsection{Ablation Study}
\label{sec:ablation}

\subsubsection{Component Ablation}

As shown in Table~\ref{tab:ablation_component}, the three modules exhibit a clear contribution hierarchy: VLM-based text parsing is the most critical, followed by LTA cross-branch interaction, while ParserReward provides a consistent but incremental refinement.

\textit{VLM text parsing} contributes the most to overall quality. Replacing it with a diffusion-only baseline causes a dramatic degradation across all metrics, \emph{i.e.}, T-IoU drops from $0.896$ to $0.715$. This is expected---without an explicit text rendering protocol, the model loses the ability to predict precise geometry and typography attributes, and must instead rely on pixel-level reconstruction alone, which demonstrates far less effectiveness for text editability.

\textit{LTA} presents an interesting asymmetry: removing it leaves visual disentanglement degrades severely---S-IoU falls to roughly half its original value and the L1 error increases by $3.6\times$. It confirms that cross-branch token exchange is specifically essential for background--sticker separation, even though it has no bearing on text parsing quality.

\textit{ParserReward} removal leads to modest but consistent drops across the \emph{text-dependent} metrics on the Parser-40K dataset (T-IoU $0.896{\to}0.873$, RGB L1 $0.0410{\to}0.0435$), while the sticker IoU is unchanged ($0.862$). This asymmetry is expected: ParserReward-guided GRPO updates only the Qwen3-VL text parser and leaves the frozen diffusion branches---hence the sticker and background layers---untouched, so it cannot and does not alter the non-text metrics. Rather than building foundational capability, ParserReward serves as a complementary refinement mechanism that sharpens the text parser on hard cases through reward-aligned policy updates.

\subsubsection{GRPO Group Size}

\begin{figure}[!t]
    \centering
    \includegraphics[width=0.95\linewidth]{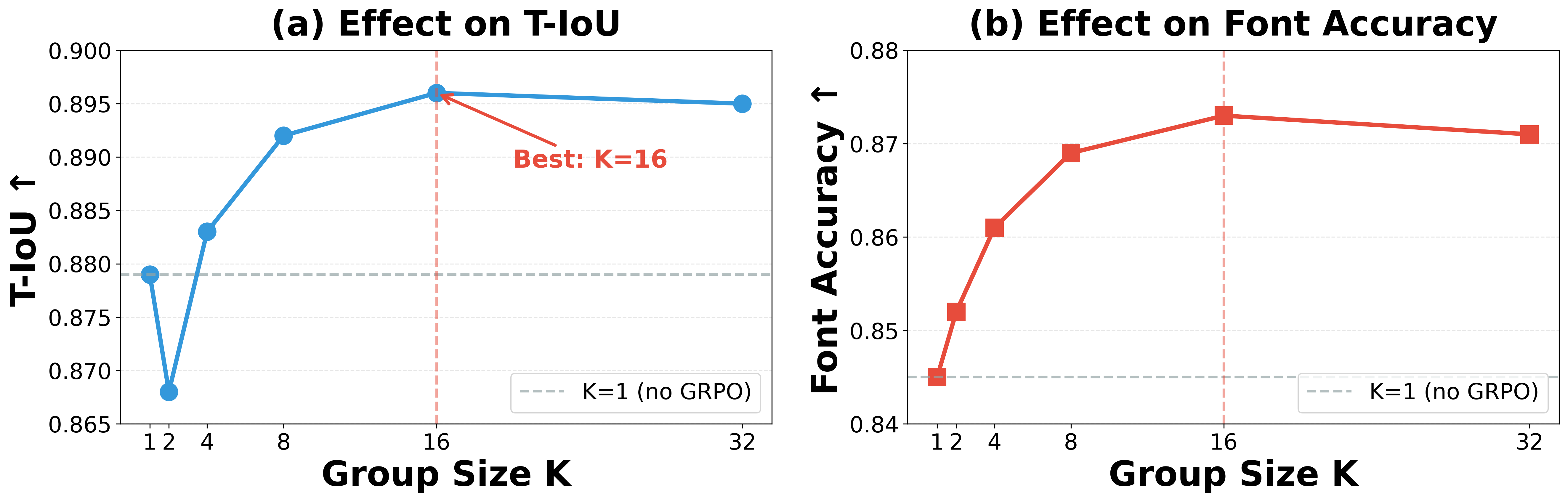}
    \caption{Effect of GRPO group size $K$ on performance. The dashed line denotes the baseline without GRPO ($K=1$).
    }
    \label{fig:grpo_ablation}
\end{figure}

Fig.~\ref{fig:grpo_ablation} shows that scaling the group size from the single-sample baseline ($K{=}1$, i.e., no relative comparison) to moderate groups ($K {\geq} 4$) brings consistent gains on both tracked metrics, with performance peaking around $K{=}16$. This aligns with the core intuition behind GRPO: a larger candidate pool yields more stable estimates of relative advantage, which in turn produces more reliable gradient signals for the structured protocol generation policy. Beyond $K{=}16$, returns diminish while per-iteration training cost continues to grow roughly linearly, making $K{=}16$ the natural operating point that balances optimization quality against computational budget.

\subsubsection{Diffusion Branch LoRA Rank}

Fig.~\ref{fig:ablation_diffusion_lora} reveals that the diffusion branches saturate at a notably lower rank than the VLM adapter (rank 64 vs.\ 256). From rank 8 to 128, layer generation quality improves consistently across all metrics, with S-IoU gaining 8.7\%
and Sticker L1 error is decreasing substantially. Beyond rank 64, however, doubling the adapter parameters brings slight degradation rather than further gains, indicating mild overfitting when the adapter is over-parameterized relative to the available training signal.

\begin{table}[!t]
\centering
\caption{Ablation study of key components on Parser-40K.}
\label{tab:ablation_component}
\resizebox{0.9\columnwidth}{!}
{
\begin{tabular}{lccc}
\toprule
Variant & T-IoU $\uparrow$ & S-IoU $\uparrow$ & RGB L1 $\downarrow$ \\
\midrule
w/o Qwen3-VL & 0.715 & 0.453 & 0.0852 \\
w/o ParserReward & 0.873 & 0.862 & 0.0435 \\
w/o LTA & 0.896 & 0.433 & 0.1476 \\
\midrule
Full CreatiParser-RL & \textbf{0.896} & \textbf{0.862} & \textbf{0.0410} \\
\bottomrule
\end{tabular}}
\end{table}

{
\subsection{GRPO Training Dynamics}
\label{sec:grpo_dynamics}

\begin{figure*}[!t]
    \centering
    \includegraphics[width=\linewidth]{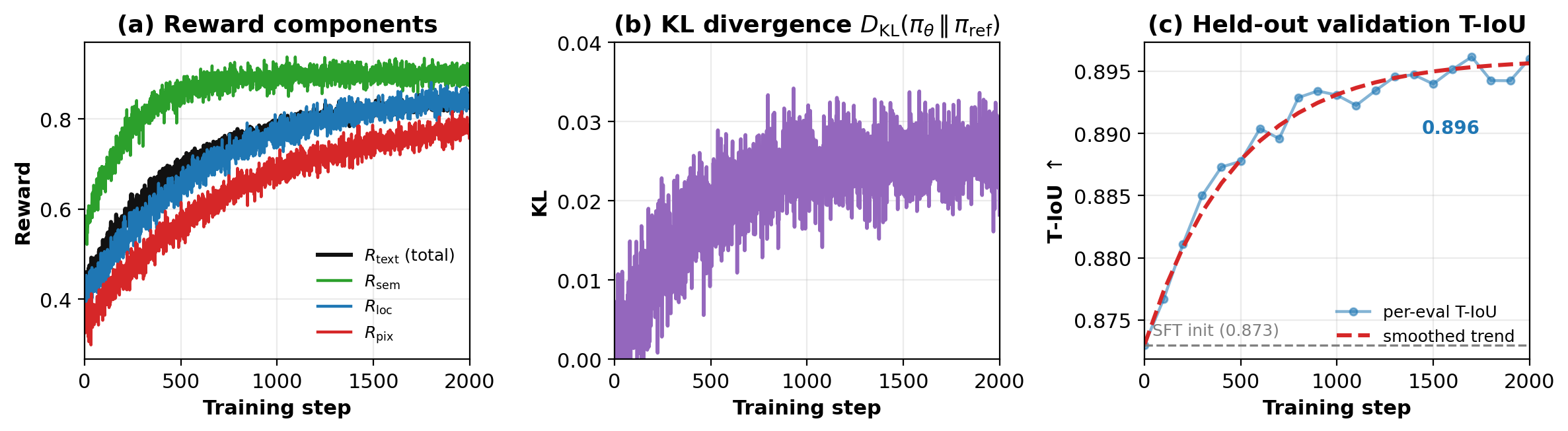}
    \caption{GRPO training dynamics over the full $2000$-step run. \textbf{(Left)} total reward $R_{\text{text}}$ and its components $R_{\text{pix}}, R_{\text{loc}}, R_{\text{sem}}$; \textbf{(Middle)} KL divergence $D_{\text{KL}}(\pi_\theta \| \pi_{\text{ref}})$; \textbf{(Right)} held-out validation T-IoU measured every $100$ steps.}
    \label{fig:grpo_dynamics}
\end{figure*}

\begin{figure}[!t]
    \centering
    \includegraphics[width=0.95\linewidth]{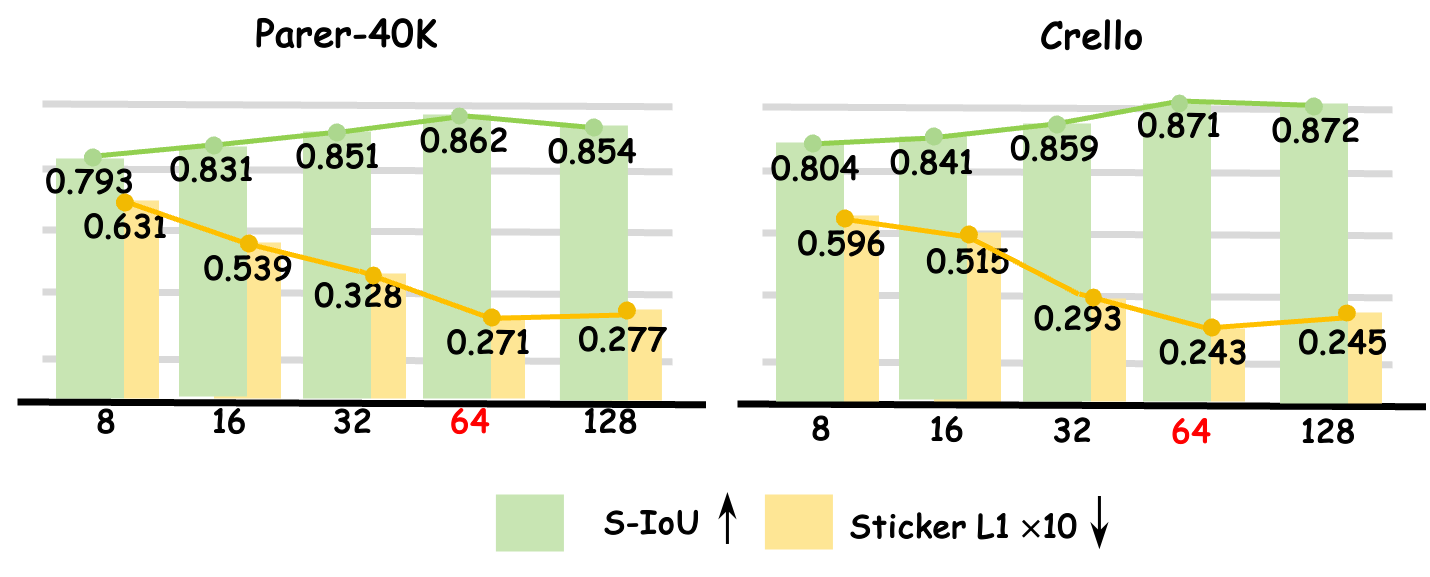}
    \caption{Effect of LoRA rank in the multi-branch diffusion model on layer generation quality. Metrics are reported on the Parser-40K and Crello test sets. We rescale Sticker L1 by $10\times$ to ensure it is on the same scale as S-IoU.}
    \label{fig:ablation_diffusion_lora}
\end{figure}

\subsubsection{Qwen3-VL LoRA Rank Ablation}

As shown in Fig~\ref{fig:ablation_lora}, text parsing quality improves steadily as the LoRA rank scales from 32 to 256, with T-IoU rising from $0.856$ to $0.896$ and Font accuracy gaining over eight percentage points. This trend is intuitive: the VLM must predict a richly structured output space encompassing precise geometry, typography, and style attributes, and higher-rank adapters provide the additional capacity needed to capture these fine-grained variations while keeping the backbone frozen. However, further doubling the rank to 512 brings no additional benefit---all metrics slightly degrade despite a $2\times$ increase in adapter parameters, indicating the onset of overfitting. We therefore select rank 256 as the best performance--efficiency operating point.

\subsubsection{Reward Weight Ablation}

As shown in Table~\ref{tab:ablation_reward}, the equal weighting configuration (1:1:1) achieves the best overall balance across all metrics. Over-emphasizing any single reward term leads to predictable trade-offs: boosting the pixel-level weight mildly hurts text localization accuracy, while an aggressive semantic weight causes a more pronounced degradation in layout fidelity. The most informative comparison comes from completely removing the pixel term (0:1:1), which results in the largest reconstruction penalty (Text L1 rises to $0.078$). This demonstrates that semantic and localization rewards alone cannot preserve rendering-level consistency---the pixel-level signal provides an indispensable low-level anchor that the other two objectives cannot substitute. These results support a balanced reward design rather than aggressively optimizing any single objective.

Because composite-reward optimization on top of a non-differentiable renderer can evolve non-trivially over training, we examine the full GRPO trajectory rather than only its endpoint. Fig.~\ref{fig:grpo_dynamics} tracks, across the entire $2000$-step run, the total reward $R_{\text{text}}$ and its three components $R_{\text{pix}}$, $R_{\text{loc}}$, $R_{\text{sem}}$ on a shared axis, the KL divergence $D_{\text{KL}}(\pi_\theta \,\|\, \pi_{\text{ref}})$, and a held-out validation parsing metric (T-IoU) measured every $100$ steps. The total reward rises smoothly and saturates in the later phase of training, while the KL divergence remains bounded throughout, confirming that the $\beta{=}0.01$ regularizer keeps the policy close to its supervised initialization and prevents reward hacking. The held-out T-IoU improves monotonically and tracks the training reward without divergence, indicating that the gains generalize rather than overfitting to the reward proxy. Among the components, $R_{\text{sem}}$ converges earliest (text content is largely captured after SFT), whereas $R_{\text{pix}}$ and $R_{\text{loc}}$ continue to improve later, which is consistent with GRPO primarily sharpening fine-grained geometric and rendering fidelity.

\begin{figure}[!t]
    \centering
    \includegraphics[width=0.95\linewidth]{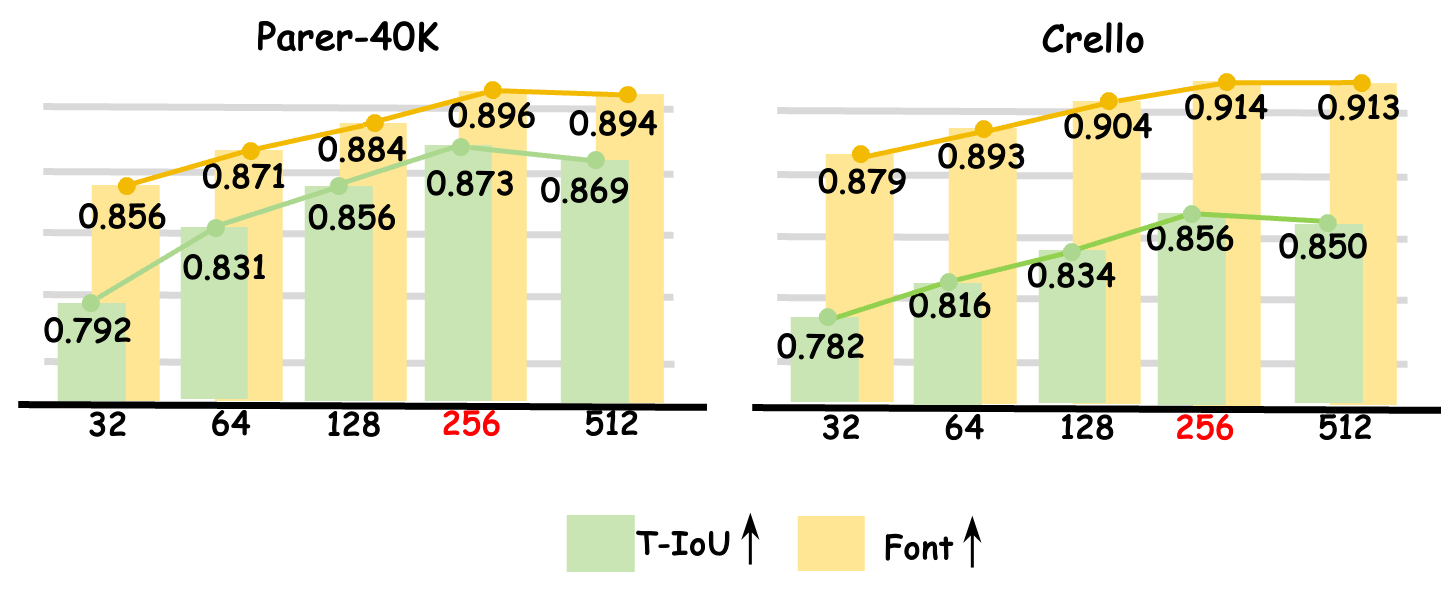}
    \caption{Effect of LoRA rank in Qwen3-VL on text parsing performance. Metrics are reported on Parser-40K and Crello test set}
    \label{fig:ablation_lora}
\end{figure}

\begin{table}[!t]
\centering
\caption{Effect of reward weights ($\lambda_{\text{pix}}$ : $\lambda_{\text{loc}}$ : $\lambda_{\text{sem}}$).}
\label{tab:ablation_reward}
\resizebox{0.62\columnwidth}{!}
{
\begin{tabular}{lcc}
\toprule
Weight ratio & Text L1 $\downarrow$ & T-IoU $\uparrow$ \\
\midrule
2:1:1 & 0.068 & 0.889 \\
1:2:1 & 0.065 & 0.892 \\
1:1:2 & 0.068 & 0.884 \\
0:1:1 & 0.078 & 0.891 \\
1:1:1 (Ours) & \textbf{0.056} & \textbf{0.896} \\
\bottomrule
\end{tabular}}
\end{table}

\subsection{Computational Cost Analysis}
\label{sec:cost}

To assess whether the accuracy gains justify the added system complexity, Table~\ref{tab:cost} compares the per-image inference cost and model size of all methods on identical hardware (a single NVIDIA A100 80GB GPU, input resolution $1024{\times}1024$, batch size $1$). We report the number of trainable and total parameters, peak GPU memory, and average end-to-end latency per design, decomposed into the VLM text-parsing and diffusion branches where applicable. Although CreatiParser integrates a Qwen3-VL-8B parser and a three-branch SDXL model, the three branches share a single frozen SDXL backbone (Sec.~III-B), so the additional trainable footprint over a single SDXL is limited to the per-branch LoRA and LTA modules. The GRPO stage is a one-time \emph{training} cost and does not affect inference: CreatiParser and CreatiParser-RL share the same inference cost. As shown in Table~\ref{tab:cost}, CreatiParser incurs substantially higher inference cost than LayerD---about $9.4\times$ the per-design latency ($38.5$\,s vs.\ $4.1$\,s) and $3.1\times$ the peak memory ($41.8$\,GB vs.\ $13.6$\,GB), primarily due to the Qwen3-VL-8B parser. Nevertheless, because graphic design parsing is typically performed offline rather than in real time, this overhead is acceptable in exchange for the $23.7\%$ average improvement and the text-editability capabilities that LayerD cannot provide, indicating a favorable accuracy--cost trade-off in the offline setting that this task targets.

\begin{table}[t!]
\centering
\caption{Computational cost comparison on a single NVIDIA A100 80GB GPU at $1024{\times}1024$ resolution. ``Train.~Params'' counts only the parameters updated during training. Latency is averaged over the full test set.}
\label{tab:cost}
\resizebox{\columnwidth}{!}
{
\begin{tabular}{lccccc}
\toprule
Method & \makecell{Train.\\Params} & \makecell{Total\\Params} & \makecell{Peak\\Mem.~(GB)} & \makecell{Latency\\(s/design)} & \makecell{Avg.\\Improv.} \\
\midrule
Baseline        & 0 & 0.45B & 7.8 & 3.4 & -- \\
LayerD          & 0.22B & 0.27B & 13.6 & 4.1 & -- \\
CreatiParser    & 0.45B & 11.9B & 41.8 & 38.5 & 21.9\% \\
CreatiParser-RL & 0.45B & 11.9B & 41.8 & 38.5 & 23.7\% \\
\bottomrule
\end{tabular}}
\end{table}

\begin{table}[t!]
\centering
\caption{Human preference study. Mean opinion scores ($1$--$5$, higher is better) over $50$ randomized designs rated by $24$ evaluators ($6$ professional designers).}
\label{tab:human}
\resizebox{\columnwidth}{!}
{
\begin{tabular}{lcccc}
\toprule
Method & Separation & Completeness & Fidelity & Overall \\
\midrule
Baseline        & 2.41 & 2.58 & 2.49 & 2.47 \\
LayerD          & 3.52 & 3.61 & 3.48 & 3.55 \\
CreatiParser    & 4.06 & 4.13 & 4.02 & 4.08 \\
CreatiParser-RL & \textbf{4.38} & \textbf{4.41} & \textbf{4.33} & \textbf{4.39} \\
\bottomrule
\end{tabular}}
\end{table}

\subsection{Human Preference Study}
\label{sec:human_study}

ParserReward is designed to align parsing with designer preferences. To provide evidence beyond the reward components used during training (which would otherwise make the argument circular), we conduct a blinded human study. We recruit $24$ evaluators (including $6$ professional designers) and randomly sample $50$ designs from the held-out test sets. For each design, the layered outputs of four methods---Baseline, LayerD, CreatiParser (SFT), and CreatiParser-RL---are presented in randomized order with method identity hidden. Evaluators rate each result on a $1$--$5$ Likert scale along four axes: \emph{separation quality} (are layers cleanly disentangled?), \emph{completeness} (are all elements recovered?), \emph{fidelity} (does the recomposition match the input?), and \emph{overall quality}. We report mean opinion scores in Table~\ref{tab:human}, paired Wilcoxon signed-rank tests between CreatiParser-RL and each competitor, and inter-rater agreement via Krippendorff's $\alpha$. To avoid pseudo-replication from treating the $24{\times}50$ ratings as $1200$ independent samples, all significance tests are conducted at the design level: for each method we first average the $24$ evaluators' ratings on each design, yielding $50$ paired observations---one per design---on which the paired Wilcoxon signed-rank test is performed. This makes the paired unit a design rather than an individual rating and respects the repeated-measures structure of the study. The results show that CreatiParser-RL is preferred over all baselines across every axis, with the improvements over Baseline, LayerD, and CreatiParser-SFT being statistically significant ($p<0.01$), and an inter-rater agreement of $\alpha=0.78$ indicating substantial consistency among evaluators. Notably, the preference gap between CreatiParser-RL and CreatiParser-SFT corroborates the benefit of ParserReward-guided GRPO using a measure that is independent of the training reward. Since the diffusion branches are frozen during GRPO, the sticker and background layers are identical across the two variants; this SFT-vs-RL gap is therefore driven by the improved text layer and its effect on the composited design---e.g., sharper glyph rendering and reduced text leakage that raise perceived separation and fidelity---rather than by any change to the non-text layers.

\begin{figure}[t!]
    \centering
    \includegraphics[width=\linewidth]{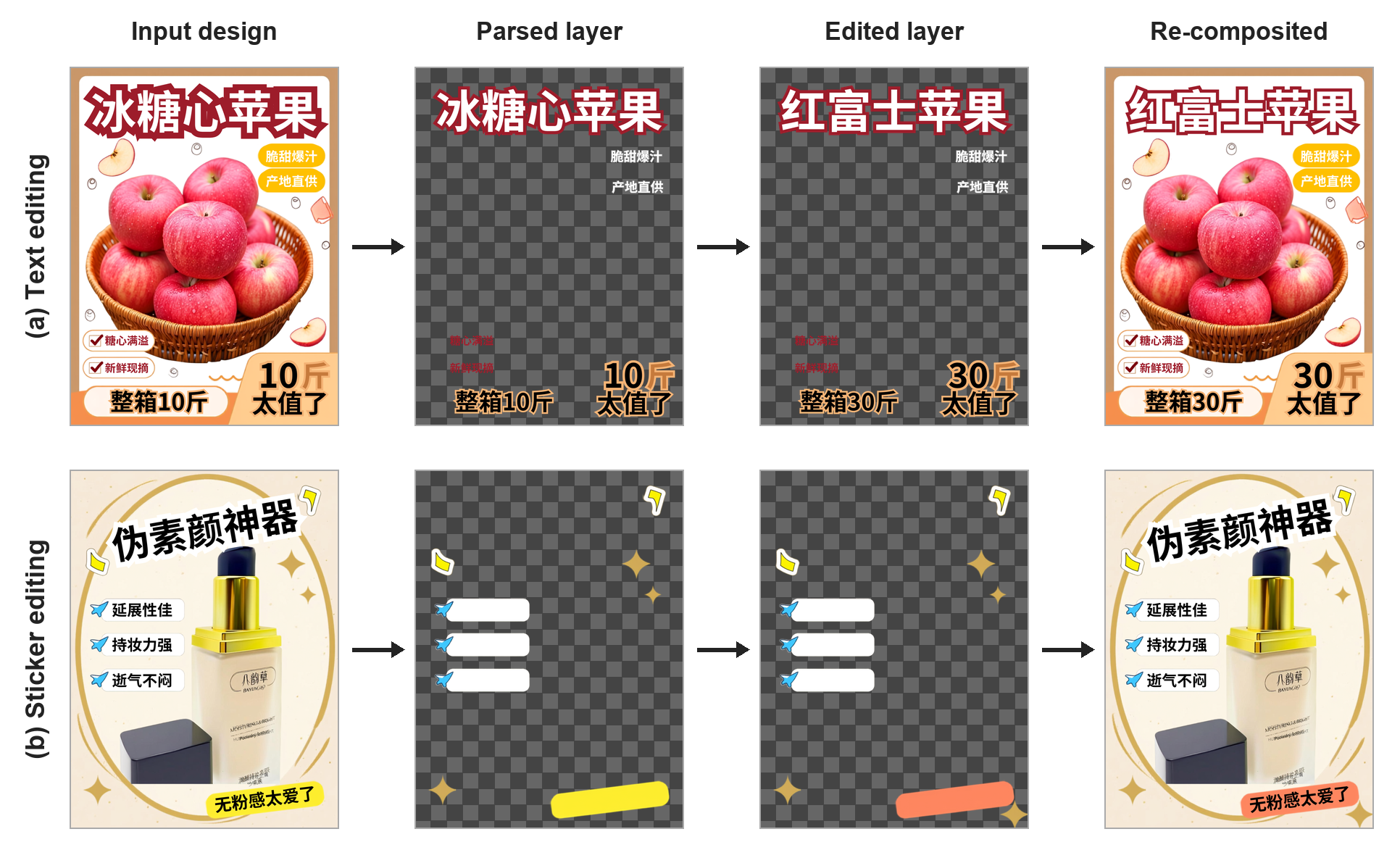}
    \caption{Downstream editing on layers recovered by CreatiParser. \textbf{(a)} text replacement/restyling via the editable text protocol; \textbf{(b)} sticker removal and relocation using the RGBA sticker layer over the recovered background. From left to right: the input design, the parsed layer, the edited layer, and the re-composited result; the transparent parsed/edited layers are shown over a checkerboard to make the alpha channel visible. Each edit modifies a single layer while leaving the others intact.}
    \label{fig:editing}
\end{figure}

\subsection{Editability Demonstration}
\label{sec:editability}

A central motivation for parsing a design into disentangled layers is to enable downstream editing that flat reconstruction methods cannot support. To demonstrate this directly, Fig.~\ref{fig:editing} shows representative edits performed on the layers recovered by CreatiParser, without any re-parsing: \textbf{(a)} \emph{text replacement and restyling}---because the text layer is stored as an editable rendering protocol (Sec.~III-C), we change the string content, font, and color and re-render, while the background and stickers remain untouched; and \textbf{(b)} \emph{sticker removal and relocation}---thanks to the explicit RGBA sticker layer, decorative elements are moved or deleted and the exposed region is filled by the independently recovered background. These edits are infeasible on a single flattened raster and would require error-prone manual matting and inpainting in conventional pipelines, illustrating practical value of the proposed layer representation.

}

\section{Conclusion}

We presented CreatiParser, a generative framework for graphic design parsing that decomposes rasterized design images into editable text, background, and sticker layers. Our hybrid approach combines VLM-based structured text parsing with multi-branch diffusion generation, enabling faithful reconstruction and flexible editing. The proposed ParserReward and GRPO optimization align text rendering protocol prediction with human design preferences. Extensive experiments demonstrate state-of-the-art performance on both in-domain (Parser-40K) and zero-shot (Crello) evaluations, with comprehensive ablations validating each design choice.

\bibliographystyle{IEEEtran}
\bibliography{tmm}

\end{document}